\documentclass[journal]{IEEEtran}

\usepackage{times}
\usepackage{epsfig}
\usepackage{graphicx}
\usepackage{subcaption}
\usepackage{amsmath}
\usepackage{amssymb}
\usepackage{color}
\usepackage{multirow}
\usepackage{bm}
\usepackage{dsfont}
\usepackage{algorithm}
\usepackage{algorithmic}
\usepackage{array}
\usepackage{textcomp}
\usepackage[normalem]{ulem}
\usepackage{etoolbox}

\graphicspath{{Figures/}}

\usepackage[breaklinks=true,bookmarks=false]{hyperref}

\newcommand{\qed}{\nobreak \ifvmode \relax \else
      \ifdim\lastskip<1.5em \hskip-\lastskip
      \hskip1.5em plus0em minus0.5em \fi \nobreak
      \vrule height0.75em width0.5em depth0.25em\fi}

\usepackage{xspace}

\newcommand{\MyMapTemplatePrefixc}[4]{\expandafter#1\csname#3#4\endcsname{#2{#4}}} % it remembles a template: \#3#4 --> #2{#4}
\forcsvlist{\MyMapTemplatePrefixc {\def} {\mathcal}{c}} {A,B,C,D,E,F,G,H,I,J,K,L,M,N,O,P,Q,R,S,T,U,V,W,X,Y,Z}  % e.g., \cA --> \mathcal{A}

\newcommand{\MyMapTemplatePrefixtb}[5]{\expandafter#1\csname#4#5\endcsname{#2{#3{#5}}}} % it remembles a template: \#3#4 --> #2{#4}
\forcsvlist{\MyMapTemplatePrefixtb {\def} {\tilde}{\mathbf}{t}} {a,b,c,d,e,f,g,h,i,j,k,l,m,n,o,p,q,r,s,t,u,v,w,x,y,z}  % e.g., \ta --> \tilde{\a}

\newcommand{\MyMapTemplateNoPrefix}[3]{\expandafter#1\csname#3\endcsname{#2{#3}}}
\forcsvlist{\MyMapTemplateNoPrefix {\def} {\mathbf}} {0, 1, a, b, c, d, e, f, g, h, i, j, k, l, m, n, o, p, q, r, u, v, w, x, y, z} % e.g., \a --> \mathbf{a}
\forcsvlist{\MyMapTemplateNoPrefix {\def} {\mathbf}} {A,B,C,D,E,F,G,H,I,J,K,L,M,N,O,P,Q,R,S,T,U,V,W,X,Y,Z}  % e.g., \A --> \mathbf{A}

\def\ie{\emph{i.e.}}
\def\eg{\emph{e.g.}}

\def\tr{\mbox{tr}}

\def\etal{\emph{et al.}\@\xspace}
\def\etc{\emph{etc}\@\xspace}
\def\ie{\emph{i.e.}\@\xspace}
\def\eg{\emph{e.g.}\@\xspace}
\def\resp{\emph{resp.}\@\xspace}
\def\wrt{\emph{w.r.t.}\@\xspace}

\begin{document}
%%%%%%%%% TITLE
\title{Zero-Shot Learning via Category-Specific Visual-Semantic Mapping}

\author{Li Niu, Jianfei Cai, and Ashok Veeraraghavan\
\thanks{L. Niu is with Electric and Computer Engineering (ECE) department in Rice University, Houston, TX 77005, USA (e-mail:ln7@rice.edu).}
\thanks{J. Cai is with the School of Computer Engineering, Nanyang Technological University, Singapore (e-mail:asjfcai@ntu.edu.sg).}
\thanks{A. Veeraraghavan is with Electric and Computer Engineering (ECE) department in Rice University, Houston, TX 77005, USA (e-mail:vashok@rice.edu).}
}

\maketitle

\begin{abstract}
Zero-Shot Learning (ZSL) aims to classify a test instance from an unseen category based on the training instances from seen categories, in which the gap between seen categories and unseen categories is generally bridged via visual-semantic mapping between the low-level visual feature space and the intermediate semantic space. However, the visual-semantic mapping (\ie, projection) learnt based on seen categories may not generalize well to unseen categories, which is known as the projection domain shift in ZSL. To address this projection domain shift issue, we propose a method named Adaptive Embedding ZSL (AEZSL) to learn an adaptive visual-semantic mapping for each unseen category, followed by progressive label refinement. Moreover, to avoid learning visual-semantic mapping for each unseen category in the large-scale classification task, we additionally propose a deep adaptive embedding model named Deep AEZSL (DAEZSL) sharing the similar idea (\ie, visual-semantic mapping should be category-specific and related to the semantic space) with AEZSL, which only needs to be trained once, but can be applied to arbitrary number of unseen categories.
Extensive experiments demonstrate that our proposed methods achieve the state-of-the-art results for image classification on three small-scale benchmark datasets and one large-scale benchmark dataset.
\end{abstract}

\section{Introduction} \label{sec:intro}

Conventional classification tasks require the training and test categories to be identical, but collecting annotated data for all categories is time-consuming and expensive in large-scale or fine-grained classification tasks. Therefore, Zero-Shot Learning (ZSL)~\cite{lampert2014attribute,guo2017zero,luo2018zero}, which aims to recognize the test instances from the categories previously unseen in the training stage, has become increasingly popular in the field of computer vision. In ZSL, the gap between seen (\ie, training) categories and unseen (\ie, testing) categories is generally bridged based on the intermediate semantic representations. Semantic representation of each category means the representation of this category in the semantic embedding space. One popular semantic representation is manually designed attribute vector ~\cite{lampert2014attribute,farhadi2009describing}, which is a high-level description for each category, such as the shape (\eg, cylindrical), material (\eg, cloth), and color (\eg, white). Besides, there also exist automatically generated semantic representations~\cite{akata2015evaluation, frome2013devise, socher2013zero, yu2013designing} including the textual features extracted from online corpus corresponding to this category or the word vector of this category name.  

Recently, based on the intermediate semantic representation (\ie, semantic embedding), many ZSL approaches have been proposed, which can be roughly categorized into Semantic Relatedness (SR) and Semantic Embedding (SE) methods according to~\cite{fu2015zero,li2015max}. In particular, the first category SR methods~\cite{mensink2014costa,norouzi2013zero,zhang2015zero} tend to learn the visual classifiers for unseen categories using the similarities between unseen categories and seen categories while the second category SE methods attract more research interest~\cite{lampert2014attribute, bucher2016improving, xu2017transductive, jayaraman2014decorrelating,kodirov2017semantic,kodirov2015unsupervised,
shojaee2016semi,frome2013devise, fu2014transductive, li2015semi, xian2016latent, zhang2016zerose, akata2015evaluation, li2015max, romera2015embarrassingly, guo2016transductive, long2016attribute, al2016recovering}, which aim to build the semantic link between visual features and semantic representations using mapping functions. More details of SR and SE methods can be found in Section~\ref{sec:related}. For Semantic Embedding (SE) methods, a mapping function needs to be learnt from the seen categories in the training stage, and then applied to the unseen categories in the testing stage. However, the visual-semantic mappings between the seen categories and unseen categories might be considerably different. In this sense, the learnt mapping (\ie, projection) may not generalize well to the test set and thus results in poor performance, leading to the recently highlighted projection domain shift, which was first mentioned in \cite{fu2014transductive} and then theoretically proved in \cite{romera2015embarrassingly}. 

To cope with the projection domain shift, some semi-supervised or transductive (the two terminologies semi-supervised and transductive are often interchangeably used in the field of ZSL) ZSL approaches have been proposed by utilizing  unlabeled test instances from unseen categories in the training stage. In general, these methods either learn a common mapping function for both seen and unseen categories, or adapt the mapping function learnt based on the seen categories to the unseen categories. Nevertheless, this may be insufficient to tackle the projection domain shift because the mapping function of each individual category varies significantly. For example, as shown in \cite{fu2014transductive}, the categories ``Zebra" and ``Pig" share the same ``hasTail" attribute, but in fact the visual appearances of their tails differ greatly. Another example will be given in our experiments, where many categories have the common attributes ``cloth" and ``cluttered space" while the visual appearances of these two attributes are dramatically diverse for these categories (see Section~\ref{sec:exp} for more details). 

To this end, we focus on the projection domain shift and tend to address this problem by learning a category-specific mapping function for each unseen category, which has never been studied before. Since we do not have labeled instances for the unseen categories, the mapping functions of unseen categories can only be learnt by transferring from those of seen categories. However, it is hard to tell which seen categories are semantically closer to a given unseen category \wrt certain entry in the semantic representation. Thus, we make a simple yet effective assumption that when the semantic representations of two categories are similar, the common non-zero entries of these two semantic representations should be semantically similar, and thus the mapping functions of these two categories should also be similar. Specifically, for each unseen category, we calculate the similarities between this unseen category and all the seen categories based on their semantic representations, and assign higher weights to the classification tasks corresponding to the more similar seen categories. This idea can be unified with many existing ZSL works with various forms of classification losses. As an instantiation, we build our method upon ESZSL~\cite{romera2015embarrassingly} considering its simplicity and effectiveness, which is named as Adaptive Embedding ZSL (AEZSL). Note that for AEZSL, we only utilize the semantic representations of unseen categories to learn the mapping functions. In order to further utilize the unlabeled instances from unseen categories like previous semi-supervised or transductive ZSL works~\cite{fu2014transductive, xu2017transductive, kodirov2015unsupervised,shojaee2016semi,li2015max}, we propose a label refinement strategy following AEZSL, which can refine the predicted test labels and update the visual classifiers for unseen categories alternatively by exploiting the relations among unseen categories and among test instances. 

One problem of AEZSL is its inefficiency for large-scale classification task with a large number of unseen categories because one visual-semantic mapping needs to be learnt for each unseen category. Thus, we aim to design a model which only needs to be trained once on seen categories but can be applied to any unseen category without retraining the model. With this aim, we develop a Deep Adaptive Embedding ZSL (DAEZSL) model, which only utilizes seen categories in the training stage but has the generalization ability to arbitrary number of unseen categories. Specifically, instead of learning one mapping for each unseen category based on its semantic representation as for AEZSL, we target at learning a projection function from semantic representation to visual-semantic mapping. In this sense, given a new unseen category, its visual-semantic mapping can be easily generated based on its semantic representation. Nevertheless, the size of visual-semantic mapping matrix is usually very large. To further simplify the task, we opt for learning the projection from semantic representation to feature mask which can be applied on visual features via elementwise product, considering that the size of feature mask is much smaller than that of visual-semantic mapping matrix. In the meantime, transforming visual features by virtue of category-specific feature masks can implicitly adapt the visual-semantic mapping to different categories. In the testing stage, given a set of unseen categories associated with semantic representations, our model is able to generate category-specific feature masks for each unseen category, which can better fit the classification tasks for unseen categories.

Our contributions are threefold: 1) this is the first work to address the projection domain shift by learning category-specific visual-semantic mappings with the idea that higher weights should be assigned to the classification tasks of more relevant seen categories for each unseen category. As an instantiation, we build our AEZSL method upon ESZSL~\cite{romera2015embarrassingly}, followed by a progressive label refinement strategy; 2) we additionally propose a deep adaptive embedding model DAEZSL for large-scale ZSL, which needs to be trained only once on seen categories but can be applied to arbitrary unseen category; 3) comprehensive experiments are conducted on three small-scale datasets and one large-scale dataset to demonstrate the effectiveness of our approaches.

%The rest of the paper is organized as follows. We discuss the related works in Section~\ref{sec:related} and briefly introduce some background knowledge in Section~\ref{sec:background}. In Section~\ref{sec:ours}, we introduce the technical details of our proposed methods including AEZSL, AEZSL\_LR, and DAEZSL. In Section~\ref{sec:exp}, we demonstrate the effectiveness of our methods with extensive experimental results including both quantitative and qualitative analyses. Finally, we conclude the paper and point out some future directions in Section~\ref{sec:conclusion}.

\section{Related Work} \label{sec:related}
As discussed in Section~\ref{sec:intro}, the existing Zero-Shot Learning (ZSL) methods can be categorized into Semantic Relatedness (SR) approaches and Semantic Embedding (SE) approaches. From another perspective, ZSL methods can be categorized into standard ZSL and transductive/semi-supervised ZSL based on whether to use the unlabeled test instances from unseen categories in the training stage. Moreover, several deep learning models have been proposed for ZSL. Additionally, since the terminology domain shift is commonly used in the field of domain adaptation, we briefly describe the difference of this terminology used in ZSL and domain adaptation.

\noindent\textbf{Semantic Relatedness (SR) and Semantic Embedding (SE) ZSL:} For SR approaches, the methods in \cite{mensink2014costa, zhang2015zero, norouzi2013zero} construct the visual classifiers for unseen categories from those for seen categories based on the semantic similarities between seen categories and unseen categories. Then, an extension was made in \cite{changpinyo2016synthesized}, which assumes that the visual classifiers for both seen and unseen categories can be represented by a set of base classifiers. The above SR approaches do not exhibit obvious projection domain shift problem, but they cannot take full advantage of the semantic representations. Moreover, the above works did not discuss how to utilize the unlabeled test instances in the training stage. In contrast, our methods can fully exploit the semantic representations and leverage the unlabeled test instances during the training procedure.

For SE approaches, three different strategies are employed to build the semantic link between the visual feature space and the semantic representation space as discussed in Section~\ref{sec:intro}. Firstly, the methods in \cite{lampert2014attribute, bucher2016improving, xu2017transductive, jayaraman2014decorrelating} are proposed for projecting the visual features to semantic representations based on the learnt attribute classifiers. Secondly, the approaches in \cite{kodirov2015unsupervised,shojaee2016semi} target at projecting the semantic representations to visual features based on the learnt dictionary. Thirdly, the works in \cite{frome2013devise, fu2014transductive, li2015semi, xian2016latent, zhang2016zerose, akata2015evaluation, li2015max, romera2015embarrassingly, guo2016transductive, long2016attribute, al2016recovering} tend to map the visual feature space and the semantic representation space into a common space, or learn a mapping function which measures the compatibility between visual features and semantic representations. Among the above SE approaches, the approaches in \cite{xian2016latent, zhang2016zerose, bucher2016improving, akata2015evaluation, lampert2014attribute, long2016attribute} do not address the projection domain shift while the remaining transductive or semi-supervised approaches can somehow account for the projection domain shift problem, which will be detailed next.

\noindent\textbf{Transductive or Semi-Supervised ZSL:} Some transductive or semi-supervised Semantic Embedding (SE) methods~\cite{fu2014transductive, xu2017transductive, kodirov2015unsupervised,shojaee2016semi,li2015max,guo2016transductive,li2015semi,rohrbach2013transfer} can alleviate the projection domain shift by utilizing the unlabeled test instances from unseen categories in the training phase. Specifically, the projection domain shift is rectified in \cite{fu2014transductive} by projecting visual features and multiple semantic embeddings of test instances into a common subspace via Canonical Correlation Analysis (CCA). 
Label propagation for zero-shot learning is used in both \cite{fu2014transductive} and \cite{rohrbach2013transfer}.
The approach in \cite{kodirov2015unsupervised} learns dictionaries for the seen categories and the unseen categories separately while enforcing these two dictionaries to be close. The methods in \cite{li2015max, shojaee2016semi, guo2016transductive, li2015semi} learn the mapping function and simultaneously infer the test labels. In \cite{li2015semi, xu2017transductive}, a Laplacian regularizer is employed based on the semantic representations or the label representations of test instances. The above transductive or semi-supervised ZSL approaches are able to account for the projection domain shift problem. However, the approach in \cite{fu2014transductive} requires multi-view semantic embeddings, which are often unavailable. Besides, the methods in \cite{kodirov2015unsupervised, li2015max, shojaee2016semi, guo2016transductive, li2015semi, xu2017transductive} either learn a common mapping function for all the categories or adapt the mapping function learnt from the seen categories to the unseen categories, which is likely to be improper for the real-world applications since the mapping function of each category may vary significantly. In contrast, our methods learn a category-specific mapping for each unseen category, which can capture diverse semantic meanings of the same attribute for different categories.

\noindent\textbf{Deep ZSL:} Compared with traditional ZSL methods based on extracted deep learning features,  there are fewer end-to-end deep ZSL models~\cite{frome2013devise, socher2013zero, yang2014unified,lei2015predicting,zhang2016learning}. These works learn the mapping from visual features to semantic representations~\cite{socher2013zero}, map visual feature space and  semantic  space  to a common space~\cite{frome2013devise,yang2014unified,zhang2016learning}, or directly learn classifiers for different categories based on their semantic representations~\cite{lei2015predicting}. However, none of the above deep ZSL methods mentions the projection domain shift issue. In contrast, we focus on tackling the projection domain shift in this paper and propose a deep adaptive embedding model DAEZSL which can adapt the visual-semantic mapping to different categories implicitly by learning category-specific feature mask. 

\noindent\textbf{Domain Adaptation:} Domain adaptation methods aim to address the domain shift issues, \ie, to reduce the domain distribution mismatch between the source domain (\ie, training set) and the target domain (\ie, test set). Regarding the domain shift problem, domain adaptation methods focus on the difference of marginal probability or class conditional probability between the source domain and the target domain while ZSL methods focus on the difference of visual-semantic mappings (\ie, projections) between seen categories and unseen categories, which is referred to as projection domain shift. Thus, the terminology of domain shift has quite different meanings in these two fields. In this paper, we focus on the projection domain shift problem in ZSL.

\section{Background} \label{sec:background}
In this paper, for ease of presentation, a vector/matrix is denoted by a lowercase/uppercase letter in boldface. The transpose of a vector/matrix is denoted using the superscript~$'$. We  use $\A\circ\B$ to denote the dot product of two matrices. Moreover, we use $\I$ to denote the identity matrix and $\A^{-1}$ to denote the inverse matrix of $\A$. We use upperscript $s$ (\resp, $t$) to indicate seen (\resp, unseen) categories while omitting the upperscript for not being specific with seen or unseen categories.

\subsection{Problem Definition} \label{sec:prob_def}
Assume we have $C^s$ seen categories and $C^t$ unseen categories. Let us denote the training (\resp, test) data from seen (\resp, unseen) categories as $\X^s\in\mathcal{R}^{d\times n^s}$ (\resp, $\X^t\in\mathcal{R}^{d\times n^t}$), where $d$ is the dimensionality of visual features and $n^s$ (\resp, $n^t$) is the number of training (\resp, test) instances  from seen (\resp, unseen) categories. We assume each category is associated with an $a$-dim semantic representation and thus the semantic representations of seen (\resp, unseen) categories can be stacked as $\A^s\in\mathcal{R}^{a\times C^s}$ (\resp, $\A^t\in\mathcal{R}^{a\times C^t}$). In order to bridge the gap between visual features and semantic representations, we develop our method using the third strategy of SE discussed in Section~\ref{sec:related}, which can exploit the discriminative capacity of semantic representations as claimed in~\cite{romera2015embarrassingly}.
Specifically, inspired by \cite{romera2015embarrassingly,guo2016transductive,qiao2016less}, we use the mapping matrix $\W\in\mathcal{R}^{d\times a}$ to match visual features with semantic representations, \ie, ${\X^s}'\W\A^s$, which measures the compatibility between instances and categories. 
Ideally, the most compatible semantic representation of each training instance should be from its ground-truth category. In the testing stage, a test instance $\x^t$ is assigned to the category corresponding to the maximum value in the $C^t$-dim vector ${\x^t}'\W\A^t$, in which $\W\A^t$ is essentially the stacked visual classifiers for unseen categories. 

%As discussed in Section~\ref{sec:intro}, in order to rectify the domain shift, we propose to learn a category-specific mapping $\W^c$ for the $c$-th unseen category based on the similarities between unseen categories and seen categories, which will be detailed in Section~\ref{sec:category_mapping}. In addition, we further adapt $\W^c$'s to the unseen categories implicitly by updating visual classifiers and refining predicted test labels alternatively, which will be fully described in Section~\ref{sec:progressive_refinement}.

\subsection{Embarrassingly Simple Zero-Shot Learning (ESZSL)}
Before introducing our method, we briefly introduce Embarrassingly Simple Zero-Shot Learning (ESZSL)~\cite{romera2015embarrassingly}, based on which we build our own method considering its simplicity and effectiveness. ESZSL is formulated as

\vspace{-15pt}
\begin{eqnarray} \label{eqn:ESZSL}
\min_{\W}&&\|{\X^s}'\W\A^s-\Y^s\|_F^2 + \gamma \|\W\A^s\|_F^2 \\
&&+\lambda \|{\X^s}'\W\|_F^2+\beta\|\W\|_F^2,
\end{eqnarray}

\noindent in which $\gamma$, $\lambda$, and $\beta$ are trade-off parameters, and $\Y^s\in\mathcal{R}^{n^s\times C^s}$ is a binary label matrix with the $i$-th row being the label vector of the $i$-th training instance. Note that in (\ref{eqn:ESZSL}),
$\|{\X^s}'\W\A^s-\Y^s\|_F^2$ can fully exploit the discriminability of semantic representations by assigning each instance to the category with the most compatible semantic representation, and $\|\W\A^s\|_F^2$ (\resp, $\|{\X^s}'\W\|_F^2$) is used to control the complexity of the projection of semantic embeddings (\resp, visual features) onto the visual feature (\resp, semantic embedding) space. By setting the derivative of (\ref{eqn:ESZSL}) \wrt $\W$ as zeros, we can have

\vspace{-15pt}
\begin{eqnarray} \label{eqn:ESZSL_sub1}
(\X^s{\X^s}'\!+\!\gamma\I)\W\A^s{\A^s}'\!+\! \lambda(\X^s{\X^s}'\!+\!\frac{\beta}{\lambda}\I)\W\!=\!\X^s\Y^s{\A^s}'.
\end{eqnarray}

\noindent when setting $\beta=\gamma\lambda$, the problem in (\ref{eqn:ESZSL_sub1}) has a close-form solution:

\vspace{-15pt}
\begin{eqnarray} \W=(\X^s{\X^s}'+\gamma\I)^{-1}\X^s\Y^s{\A^s}'
(\A^s{\A^s}'+\lambda\I)^{-1}.
\end{eqnarray}

\section{Our Methods} \label{sec:ours}
In this section, we first build our AEZSL method upon ESZSL to learn category-specific mapping matrix in Section~\ref{sec:category_mapping} followed by a label refinement strategy in Section~\ref{sec:progressive_refinement}. Moreover, we propose a deep adaptive embedding model named DAEZSL for large-scale ZSL in Section~\ref{sec:DAEZSL}, which shares the similar idea with AEZSL.

\subsection{Adaptive Embedding Zero-Shot Learning}
In this section, we first introduce how to learn category-specific visual-semantic mapping, followed by describing our label refinement strategy.

\subsubsection{Category-Specific Visual-Semantic Mapping} \label{sec:category_mapping}
Since the visual-semantic mappings of different categories could be largely different, we aim to learn a category-specific mapping matrix for each unseen category (\ie, $\W^c$ for the $c$-th unseen category) to address the projection domain shift. Because no labeled instances are provided for the unseen categories, we can only transfer from the mapping matrices of seen categories. However, it is a challenging task to determine which seen categories are more semantically similar to a given unseen category \wrt certain entry in the semantic representation. To facilitate the transfer, we assume that when the semantic representations of two categories are similar, the common non-zero entries shared by these two semantic representations should be semantically similar, and thus the mapping matrices of these two categories should also be similar. 

Specifically, recall that in (\ref{eqn:ESZSL}), the $\tilde{c}$-th column of ${\X^s}'\W\A^s-\Y^s$ corresponds to the classification task for the $\tilde{c}$-th  seen category, and the tasks for different seen categories are dependent on one another with a common $\W$. Given the $c$-th unseen category, in order to ensure that $\W^c$ is close to the mapping matrices of those similar seen categories, we assign higher weights to the classification tasks for those more similar seen categories. In particular, we formulate this idea as $({\X^s}'\W^c\A^s-\Y^s)\S^c$, where $\S^c\in\mathcal{R}^{C^s\times C^s}$ is a diagonal matrix with the $\tilde{c}$-th diagonal element being the cosine similarity $s^c_{\tilde{c}}=\frac{{\a^t_c}'\a^s_{\tilde{c}}}{\|\a^t_c\|\|\a^s_{\tilde{c}}\|}$, which measures the similarity between the semantic representation $\a^t_c$ of the $c$-th unseen category and the semantic representation $\a^s_{\tilde{c}}$ of the $\tilde{c}$-th seen category.
Note that  $({\X^s}'\W^c\A^s-\Y^s)\S^c$ is equivalent to multiplying the $\tilde{c}$-th column of ${\X^s}'\W^c\A^s-\Y^s$ by $s^c_{\tilde{c}}$. For better explanation, assuming that the $c$-th unseen category is similar to the $\tilde{c}$-th seen category and their semantic representations share the common non-zero entry indices $\{j_1,j_2,\ldots,j_l\}$, then a higher weight $s^c_{\tilde{c}}$ should be assigned to ${\X^s}'\W^c\a^s_{\tilde{c}}-\y^s_{\tilde{c}}$ with $\y^s_{\tilde{c}}$ being the $\tilde{c}$-th column of $\Y^s$. In this way, the learnt $\W^c$ should be closer to the mapping matrix of the $\tilde{c}$-th seen category \wrt the $\{j_1,j_2,\ldots,j_l\}$-th columns.
To this end, we tend to solve $\W^c$'s for all unseen categories simultaneously using the following formulation:

\vspace{-15pt}
\begin{eqnarray} \label{eqn:ESZSL_category}
\min_{\W^c} \!\!\!\!\!\!\!\!\!\!\!\!&&\frac{1}{2}\sum_{c=1}^{C^t}\|({\X^s}'\W^c\A^s-\Y^s)\S^c\|_F^2 + \frac{\lambda_1}{2}\sum_{c=1}^{C^t}\|{\X^s}'\W^c\|_F^2\nonumber\\
&& + \frac{\lambda_2}{2}\sum_{c=1}^{C^t}\|\W^c\|_F^2+\frac{\lambda_3}{2}\sum_{c<\tilde{c}}\|\W^c-\W^{\tilde{c}}\|_F^2,
\end{eqnarray}

\noindent where $\lambda_1$, $\lambda_2$, and $\lambda_3$ are trade-off parameters, which can be obtained using cross-validation, and $\sum_{c<\tilde{c}}\|\W^c-\W^{\tilde{c}}\|_F^2$ is a co-regularizer which encourages $\W^c$'s for different unseen categories to share some common parts. Note that we omit the regularizer $\|\W^c\A^s\|_F^2$ in (\ref{eqn:ESZSL}) for ease of optimization. When $\lambda_3$ approaches Infinity, the mappings of all categories are enforced to be the same $\W$. In this case, the problem in (\ref{eqn:ESZSL_category}) reduces to

\noindent
\begin{eqnarray}
\!\!\!\!\!\!\min_{\W} \!\!\!\!\!\!\!\!\!\!\!\!&&\frac{1}{2}\|({\X^s}'\W\A^s\!\!-\!\!\Y^s)\tilde{\S}^c\|_F^2 \!+\! \frac{\lambda_1}{2}\|{\X^s}'\W\|_F^2\!+\! \frac{\lambda_2}{2}\|\W\|_F^2,\nonumber
\end{eqnarray}
in which the $\tilde{\S}^c$ is a diagonal matrix with the $\tilde{c}$-th diagonal element being $\sqrt{\frac{\sum_{c=1}^{C^t} ({s_{\tilde{c}}^c)}^2}{C^t}}$.
Compared with (\ref{eqn:ESZSL}), we assign higher weights on the classification tasks corresponding to the seen categories which are closer to the overall unseen categories based on $\tilde{\S}^c$.

To solve the problem in (\ref{eqn:ESZSL_category}), we update each $\W^c$ by fixing all the other $\W^{\tilde{c}}$'s for $\tilde{c}\neq c$ in an alternating fashion. Specifically, by setting the derivative of (\ref{eqn:ESZSL_category}) \wrt $\W^c$ to zeros, we obtain the following equation:

\vspace{-15pt}
\begin{eqnarray}\label{eqn:ESZSL_category_wc}
&&\!\!\!\!\!\!\!\!\!\!\!\!(\X^s{\X^s}')\W^c(\A^s\S^c{\S^c}'{\A^s}'\!+\!\lambda_1\I) +((C^t\!-\!1)\lambda_3\!+\!\lambda_2)\W^c \nonumber\\
= &&\!\!\!\!\!\!\!\!\!\!\!\!\X^s\Y^s\S^c{\S^c}'{\A^s}'+\lambda_3\sum_{\tilde{c}\neq c}\W^{\tilde{c}}.
\end{eqnarray}

\noindent By denoting $\L=\X^s{\X^s}'$, $\T=\A^s\S^c{\S^c}'{\A^s}'+\lambda_1\I$, $\N=\X^s\Y^s\S^c{\S^c}'{\A^s}'+\lambda_3\sum_{\tilde{c}\neq c}\W^{\tilde{c}}$, and $\mu =(C^t-1)\lambda_3+\lambda_2 $, the problem in (\ref{eqn:ESZSL_category_wc}) becomes a special case of Sylvester equation \wrt $\W^c$ as follows,

\vspace{-15pt}
\begin{eqnarray} \label{eqn:W_derive2}
\L\W^c\T +\mu\W^c= \N.
\end{eqnarray}

\noindent Since $\L$ and $\T$ are symmetric real matrices, the problem in (\ref{eqn:W_derive2}) has an efficient solution. Inspired by \cite{simoncini2016computational}, we perform Singular Value Decomposition (SVD) on $\L$ and $\T$, \ie, $\L=\hat{\U}\bm{\Sigma}^L\hat{\U}'$ and $\T=\hat{\V}\bm{\Sigma}^T\hat{\V}'$ with $\bm{\Sigma}^L$ (\resp, $\bm{\Sigma}^T$) being a diagonal matrix,  in which the $i$-th diagonal element is $\sigma^L_i$ (\resp, $\sigma^T_i$).  Then, we can rewrite (\ref{eqn:W_derive2}) as

\vspace{-15pt}
\begin{eqnarray} \label{eqn:W_lya}
\hat{\U}\bm{\Sigma}^L\hat{\U}'\W^c\hat{\V}\bm{\Sigma}^T\hat{\V}'+\mu\W^c=\N.
\end{eqnarray}

\noindent After multiplying (\ref{eqn:W_lya}) by $\hat{\U}'$ (\resp, $\hat{\V}$) on the left (\resp, right) and considering the orthonormality of $\hat{\U}$ (\resp, $\hat{\V}$), we have

\vspace{-15pt}
\begin{eqnarray}
\bm{\Sigma}^L\hat{\U}'\W^c\hat{\V}\bm{\Sigma}^T+
\mu\hat{\U}'\W^c\hat{\V}=\hat{\U}'\N\hat{\V}.
\end{eqnarray}

By denoting $\hat{\W}^c=\hat{\U}'\W^c\hat{\V}$ and $\hat{\N}=\hat{\U}'\N\hat{\V}$, we can arrive at
\begin{eqnarray} \label{eqn:W_lya3}
\bm{\Sigma}^L\hat{\W}^c\bm{\Sigma}^T+\mu\hat{\W}^c=\hat{\N}.
\end{eqnarray}

Based on (\ref{eqn:W_lya3}), we can easily solve each entry in $\hat{\W}^c$ as $\hat{W}^c_{ij}=\frac{\hat{N}_{ij}}{\sigma^L_i \sigma^T_j+\mu}$. Note that $\L$ is positive semi-definite, $\T$ is positive definite, and $\mu>0$, so $\sigma^L_i\sigma^T_j+\mu\neq 0,\,\,\forall i,j$, and the problem in (\ref{eqn:W_lya3}) has unique solution. Then, we can recover $\W^c$ by using $\W^c=\hat{\U}\hat{\W}^c\hat{\V}'$. Because SVD on $\L$ and $\T$ can be precomputed, our algorithm is quite efficient.
We update each $\W^c$ alternatively until the objective of (\ref{eqn:ESZSL_category}) converges. We name this method as Adaptive Embedding ZSL (AEZSL) and
the algorithm to solve (\ref{eqn:ESZSL_category}) is listed in Algorithm~\ref{alg:AEZSL}. 

\setlength{\textfloatsep}{5pt}
\begin{algorithm}[t]
   \caption{The algorithm to solve AEZSL (\ref{eqn:ESZSL_category})}
   \label{alg:AEZSL}
\begin{algorithmic}[1]
   \STATE {\bfseries Input:} $\X^s, \Y^s, \A^s,\S^c, \lambda_1, \lambda_2, \lambda_3$
   \STATE Initialize all $\W^c$'s equally using ESZSL~\cite{romera2015embarrassingly}.
   \REPEAT
   	  \FOR{ $c = 1 : C^t$ }
      \STATE Update $\W^c$ by solving (\ref{eqn:W_derive2}).
      \ENDFOR
   \UNTIL The objective of (\ref{eqn:ESZSL_category}) converges.      
   \STATE {\bfseries Output:} $\W^c$'s.
\end{algorithmic}
\label{alg:RKLRR}
\end{algorithm}

After learning $\W^c$'s,  we can obtain the visual classifier for the $c$-th unseen category as $\p^c=\W^c\a^t_c$. In the testing stage, given a test instance $\x^t$ from unseen categories, we can obtain its decision value for the $c$-th unseen category as ${\p^c}'\x^t$.

\noindent\textbf{Discussion: }The projection domain shift problem has been theoretically proved in \cite{romera2015embarrassingly}, in which the seen (\resp, useen) categories are referred to as the source (\resp, target) domain. Then, two extreme cases are presented in \cite{romera2015embarrassingly} when the semantic representations of two domains are identical or orthogonal. Specifically, when the semantic representations of two domains are identical (\ie, $\a^t_c=\a^s_{\tilde{c}}$), the error bound of the learnt classifier is approximate to that of a standard classifier without projection domain shift. In this case, we have $s^c_{\tilde{c}}=1$, which allows the maximum transfer. When their semantic representations are orthogonal (\ie, ${\a^t_c}'\a^s_{\tilde{c}}=0$), the error bound is vacuous and no transfer can be done. In this case, we have $s^c_{\tilde{c}}=0$, which means no transfer. So the analysis for our problem accords with that in \cite{romera2015embarrassingly}, which verifies that it is reasonable to control how much to transfer from seen categories based on the cosine similarities.

In this paper, we build our AEZSL method upon ESZSL due to its simplicity and effectiveness. However, it is worth mentioning that the idea of AEZSL, \ie, assigning higher weights on the classification tasks of more similar seen categories for each unseen category, can be incorporated into many existing ZSL frameworks with slight modification based on their learning paradigms and used losses. 

%Now we consider the above two extreme cases in our problem. Specifically, when the semantic representations of the $c$-th unseen category and the $\tilde{c}$-th seen category are identical (\ie, $\a^t_c=\a^s_{\tilde{c}}$), we have $s^c_{\tilde{c}}=1$, which allows the maximum transfer. When these two semantic representations are orthogonal (\ie, ${\a^t_c}'\a^s_{\tilde{c}}=0$), we have $s^c_{\tilde{c}}=0$, which means no transfer. So the analysis for our problem accords with that in \cite{romera2015embarrassingly}, which verifies that it is a reasonable choice to control how much to transfer from seen categories based on the cosine similarities.

\subsubsection{Progressive Label Refinement} \label{sec:progressive_refinement}
Note that in Section \ref{sec:category_mapping}, we only utilize the semantic representations of  unseen categories to learn the adaptive mapping matrices. In order to further adapt the mapping matrices to the unseen categories by utilizing the unlabeled test instances, we propose a progressive approach to update visual classifiers and refine predicted test labels alternatively, similar to some progressive semi-supervised learning approaches like~\cite{blum1998combining}. Unlike traditional semi-supervised learning which requires the labeled and unlabeled instances to be from the same set of categories, zero-shot learning does not have any labeled instances from the unseen categories. So we divide the test set into a confident set $\mathcal{L}$ and an unconfident set $\mathcal{U}$, in which the labels in $\mathcal{L}$ (\ie, $\Y^l$) are expected to be relatively more accurate than those in $\mathcal{U}$ (\ie, $\Y^u$). Note that $\Y^l$ and $\Y^u$ are both binary label matrices, similar to $\Y^s$ in (\ref{eqn:ESZSL_category}). Initially, $\mathcal{L}$ is an empty set and $\mathcal{U}$ is the entire test set with initial labels predicted by the initial classifiers $\p^c$'s, which are obtained based on $\W^c$'s from Section~\ref{sec:category_mapping}. Then, we move $k$ most confident instances from $\mathcal{U}$ into $\mathcal{L}$, and update the visual classifiers based on the new confident and unconfident sets. We repeat the above process iteratively until $\mathcal{U}$ becomes empty and output $\Y^l$ as final predicted test labels. The details of each iteration will be described as follows, in which we stack $\p^c$'s for all unseen categories as $\P$, and split $\X^t$ into $\X^l$ and $\X^u$ corresponding to $\mathcal{L}$ and $\mathcal{U}$.

In each iteration, we first use the visual classifiers $\P$ from the previous iteration to select  $k$ most confident instances from $\mathcal{U}$ based on the confidence score, which is defined as a soft-max function $\textnormal{conf}(\x^t) = \frac{\exp({\p^{c(\x^t)}}'\x^t)}{\sum_{\hat{c}}\exp({\p^{\hat{c}}}'\x^t)}$ with $c(\x^t)$ being the assigned label of $\x^t$ corresponding to $\p^{c(\x^t)}$ which achieves the highest prediction score on $\x^t$. Note that the labels of the selected $k$ instances are updated as $c(\x^t)$ while the labels of the remaining instances in $\mathcal{U}$ stay unchanged.  

After moving the $k$ most confident instances from $\mathcal{U}$ into $\mathcal{L}$, we update $\P$ by changing some predicted labels in $\mathcal{U}$ (\ie, ${\X^u}'\P$) while using $\mathcal{L}$ as weak supervision. In particular, we tend to keep the predicted labels of confident instances unchanged while changing the predicted labels of some unconfident instances selected by a group-lasso regularizer~\cite{yuan2006model} $\|{\X^u}'\P-\Y^u\|_{2,1}$, in which the rows with non-zero entries correspond to the selected instances. We change the predicted labels of selected instances based on the similarities among unconfident instances and the similarities among unseen categories, which will be explained as follows. 1) For the similarities among unconfident instances, we employ a standard Laplacian regularizer based on the smoothness assumption that the predicted label vectors of two unconfident instances should be close to each other when their visual features are similar. 2) For the similarities among unseen categories, we construct a transition matrix $\hat{\S}$ to characterize the probabilities that one category label is changed to another, in which $\hat{S}_{i,j}$ is the cosine similarity $\frac{{\a^t_i}'\a^t_j}{\|\a^t_i\|\|\a^t_j\|}$, similar to that in Section~\ref{sec:category_mapping}. With the transition matrix $\hat{\S}$, we employ a coherent regularizer $\tr(\Y^u\hat{\S}\P'\X^u)$, which enforces the predicted labels $\P'\X^u$ to be coherent with the expected transited labels $\Y^u\hat{\S}$ based on the transition probabilities. Note that we set the diagonal elements of $\hat{\S}$ as zeros to encourage the labels to be changed. To this end, the formulation to update $\P$ in each iteration can be written as 

\vspace{-15pt}
\begin{eqnarray} \label{eqn:EZSL_refine}
\min_{\P}\!\!\!\!\!\!\!\! && \frac{1}{2}\|{\X^l}'\P-\Y^l\|_F^2  +\frac{\gamma_1}{2}\|{\X^u}'\P-\Y^u\|_{2,1}\nonumber\\
&&\!\!\!\!\!\!\!\!-\gamma_2\tr(\Y^u\hat{\S}\P'\X^u)+\frac{\gamma_3}{2}\tr(\P'\X^u\H^u{\X^u}'\P),
\end{eqnarray}

\noindent where $\gamma_1$, $\gamma_2$, and $\gamma_3$ are trade-off parameters, which can be obtained using cross-validation, and $\H^u$ is the Laplacian matrix constructed based on $\X^u$ following \cite{li2015semi}.

\setlength{\textfloatsep}{3pt}
\begin{algorithm}[t]
   \caption{The algorithm of progressive label refinement}
   \label{alg:PRZSL}
\begin{algorithmic}[1]
   \STATE {\bfseries Input:} $\X^u, \Y^u, \hat{\S}, k, \gamma_1, \gamma_2, \gamma_3$
   \STATE Initialize $\mathcal{L}$ as $\emptyset$ and $\mathcal{U}$ as the entire test set. Initialize $\P$ based on the learnt $\W^c$'s from Algorithm~\ref{alg:AEZSL}.
   \REPEAT
      \STATE Move $k$ most confident instances selected by $\P$ from $\mathcal{U}$ to $\mathcal{L}$. Update $\X^l$, $\X^u$, $\Y^l$, $\Y^u$, $\H^u$ accordingly.
   	  \REPEAT
   	  \STATE Update $\D$ based on its definition below (\ref{eqn:EZSL_refine_D}).
      \STATE Update $\P$ by using (\ref{eqn:solutionP}).      
      \UNTIL The objective of (\ref{eqn:EZSL_refine}) converges.
   \UNTIL The unconfident set $\mathcal{U}$ is empty. \\        
   \STATE {\bfseries Output:} $\Y^l$, $\P$.
\end{algorithmic}
\label{alg:RKLRR}
\end{algorithm}

The problem in (\ref{eqn:EZSL_refine}) is not easy to solve due to the regularizer $\|{\X^u}'\P-\Y^u\|_{2,1}$. According to \cite{nie2010efficient}, solving (\ref{eqn:EZSL_refine}) is equivalent to solving the following problem iteratively:

\vspace{-15pt}
\begin{eqnarray} \label{eqn:EZSL_refine_D}
\min_{\P} \!\!\!\!\!\!\!\!\!\!\!\!&&\frac{1}{2}\|{\X^l}'\P\!-\!\Y^l\|_F^2 \!+\!\frac{\gamma_1}{2}\tr((\P'\X^u\!-\!{\Y^u}')\D({\X^u}'\P\!-\!\Y^u))\nonumber\\
&&-\gamma_2\tr(\Y^u\hat{\S}\P'\X^u)+\frac{\gamma_3}{2}\tr(\P'\X^u\H^u{\X^u}'\P),
\end{eqnarray}
\noindent where $\D$ is a diagonal matrix with the $i$-th diagonal element being $\frac{1}{2\|\q_i\|_2}$, in which $\q_i$ is the $i$-th row of $\Q$ with $\Q={\X^u}'\P-\Y^u$. In each iteration of solving $\P$, we calculate $\D$ based on previous $\P$ and then update $\P$ by solving (\ref{eqn:EZSL_refine_D}). As claimed in \cite{nie2010efficient}, the updated $\P$ in each iteration will decrease the objective of (\ref{eqn:EZSL_refine}).

By setting the derivative of (\ref{eqn:EZSL_refine_D}) \wrt $\P$ to zeros, we can obtain the close-form solution to $\P$ as
\begin{eqnarray}\label{eqn:solutionP}
\P =\!\!\!\!\!&&(\X^l{\X^l}'+\gamma_1\X^u\D{\X^u}'+\gamma_3\X^u\H^u{\X^u}'+\nu\I)^{-1}\nonumber\\
&&(\X^l\Y^l+\gamma_1\X^u\D\Y^u+\gamma_2\X^u\Y^u\hat{\S}),
\end{eqnarray}
\noindent in which $\nu$ is a very small number (\ie, $\nu\rightarrow0$). We update $\P$ by solving (\ref{eqn:EZSL_refine_D}) iteratively until the objective of (\ref{eqn:EZSL_refine}) converges. According to \cite{nie2010efficient}, the output $\P$ is the global optimum solution to (\ref{eqn:EZSL_refine}) because $\frac{1}{2}\|{\X^l}'\P-\Y^l\|_F^2 -\gamma_2\tr(\Y^u\hat{\S}\P'\X^u)+\frac{\gamma_3}{2}\tr(\P'\X^u\H^u{\X^u}'\P)$ is convex \wrt $\P$.
The whole algorithm of the proposed progressive label refinement is summarized in Algorithm \ref{alg:PRZSL}. We refer to our AEZSL with Label Refinement as AEZSL\_LR. 

\noindent\textbf{Discussion: }Since $\p^c=\W^c\a^t_c$ (see Section~\ref{sec:category_mapping}) and $\a^t_c$'s are fixed, updating $\p^c$'s implies updating $\W^c$. In other words, by updating $\p^c$ based on the similarities among test instances and among unseen categories, we actually adapt the mapping matrices $\W^c$'s to the test set implicitly.

\setlength{\textfloatsep}{0pt}
\begin{figure*}[t]
        \centering
        \includegraphics[width=0.8\textwidth]{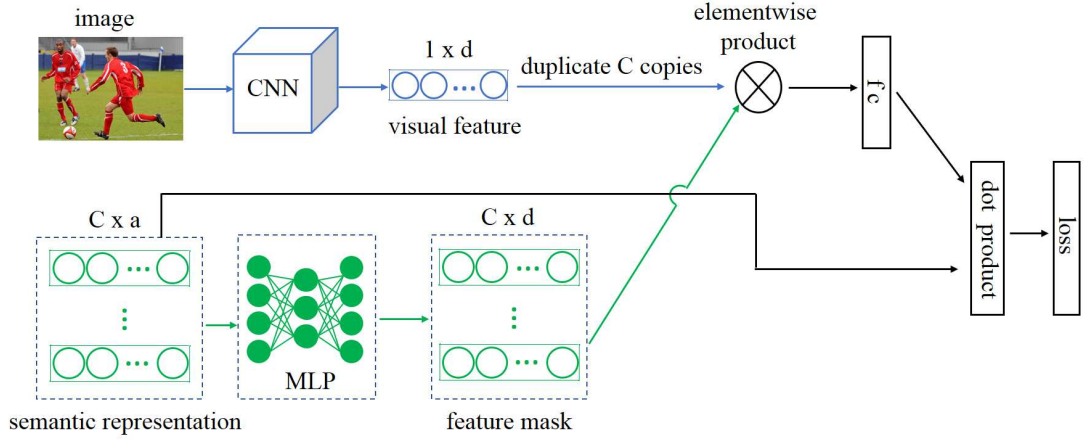}
        \caption{Deep adaptive embedding model for zero-shot learning. In the top flow, the feature of each input image is duplicated to $C$ copies. In the bottom flow,  the semantic representations of all $C$ categories pass through multi-layer perceptrons (MLP) and generate the feature masks for $C$ categories, which are applied on the duplicated features via elementwise product. Then, the masked features are fed into a fully connected (fc) layer (\ie, visual-semantic mapping), followed by dot product with semantic representations, and finally output the decision value matrix, based on which we minimize the training loss in the training stage and predict test instances in the testing stage.}
        \label{fig:DAEZSL}
        \vspace{-10pt}
\end{figure*}

\subsection{Deep Adaptive Embedding Zero-Shot Learning} \label{sec:DAEZSL}

One of the important applications of zero-shot learning is large-scale classification since it is difficult to obtain sufficient well-labeled training data exhaustively for all the categories.
However, our AEZSL method is not very efficient in this scenario, because we need to learn one visual-semantic mapping for each unseen category (\eg, over $20,000$ test categories in the ImageNet 2011 21K dataset). This concern motivates us to design a model which is more suitable for large-scale ZSL. In order to mitigate the burden induced by a large set of unseen categories, we design a deep adaptive embedding model named DAEZSL, which only needs to be trained once but can generalize to arbitrary number of unseen categories.

Similar as in Section~\ref{sec:category_mapping}, we assume the visual-semantic mappings should be category-specific and related to the semantic representation of each category. To avoid learning one mapping for each unseen category as in Section~\ref{sec:category_mapping}, one possible approach is learning a projection $f(\cdot)$ from semantic representation $\a_c$ to visual-semantic mapping $\W_c$, \ie, $f(\a_c)=\W_c$, so that given a new unseen category with semantic representation $\a_{\tilde{c}}$, we can easily obtain its visual-semantic mapping $\W_{\tilde{c}}$ by using $\W_{\tilde{c}}=f(\a_{\tilde{c}})$.

However, the size of $\W_c$ is usually very large (\ie, $d\times a$), so we adopt an alternative approach for the sake of computational efficiency, that is, learning the projection $g(\cdot)$ from semantic representation $\a_c$ to feature mask $\m_c$, \ie, $g(\a_c)=\m_c$. The feature mask $\m_c$ is applied on the visual feature via elementwise product, and thus has the same dimension $d$ as visual feature. Since the size of $\m_c$ (\ie, $d$) is much smaller than that of $\W_c$ (\ie, $d\times a$), it is much more efficient to learn the projection $g(\cdot)$ instead of $f(\cdot)$. In fact, applying the feature masks of different categories can be treated as implicitly adapting the visual-semantic mapping to different categories, which will be explained as follows. Assume we have a common visual-semantic mapping $\W$, given the $i$-th training instance with visual feature $\x_i^s$, its decision values after applying the feature mask $\m_c^s$ is
\begin{eqnarray} \label{eqn:W_adaptation}
(\x_i^{s'}\circ{\m_c^s}')\W\A^s= \x_i^{s'}(\bar{\M}_c^s\circ \W)\A^s,
\end{eqnarray}
in which $\bar{\M}_c^s$ is horizontally stacked $a$ copies of $\m_c^s$. Then, we define the implicit visual-semantic mapping $\bar{\W}_c^s$ as
\begin{eqnarray} \label{eqn:w_c_s}
\bar{\W}_c^s=\bar{\M}_c^s\circ \W,
\end{eqnarray}
from which we can see that learning feature mask $\m_c^s$ is equivalent to adapting $\W$ to $\bar{\W}_c^s$ by using the mask $\bar{\M}_c^s$.  Note that we simplify the task of adapting $\W$ by using the mask $\bar{\M}_c^s$ with all columns being the same $\m_c^s$, so that the number of variables (\ie, size of $\m_c^s$) generated by $g(\cdot)$ is greatly reduced compared with the size of visual-semantic mapping.

In practice, we only need to learn $\W$ and $g(\cdot)$ without explicitly producing $\bar{\W}_c^s$. In the remainder of this section, we use implicit $\bar{\W}_c^s$  merely for the purpose of better explanation. Specifically, we expect implicit $\bar{\W}_c^s$ to satisfy two properties: generalizability and specificity, analogous to AEZSL in (\ref{eqn:ESZSL_category}) (category-specific $\W^c$'s for specificity and co-regularizer for generalizability).

\noindent\textbf{Generalizability: }On one hand, we expect any $\bar{\W}_c^s$ can correctly classify the training instances from any category, which can be achieved by minimizing the square loss $\sum_{c=1}^{C^s}\|{\X^s}'\bar{\W}_c^s\A^s-\Y^s\|_F^2$ with $\X^s$ and $\Y^s$ being the same as defined in (\ref{eqn:ESZSL_category}).
After defining the one-hot label vector of $\x_i^s$ as $\y_i^s$ with the $c(i)$-th entry being $1$ and $\M^s\in\mathcal{R}^{d\times C^s}$ as the aggregated feature masks over all $C^s$ categories with the $c$-th column being $\m_c^s$, we can have

\vspace{-15pt}
\begin{eqnarray} \label{eqn:W_category_equal}
&&\sum_{c=1}^{C^s}\|{\X^s}'\bar{\W}_c^s\A^s-\Y^s\|_F^2,\\
=&&\sum_{c=1}^{C^s}\sum_{i=1}^{n^s} \|{\x_i^s}'\bar{\W}_c^s\A^s-\y_i^s\|^2\nonumber\\
= && \sum_{i=1}^{n^s}\sum_{c=1}^{C^s} \|{\x_i^s}' (\bar{\M}_c^s\circ\W)\A^s-\y_i^s\|^2\nonumber\\
= && \sum_{i=1}^{n^s}\sum_{c=1}^{C^s} \|(\x_i^{s'}\circ\m_c^{s'}) \W\A^s-\y_i^s\|^2\nonumber\\
= && \sum_{i=1}^{n^s}\|({\bar{\X}_i^{s'}}\circ \M^{s'})\W\A^s-\bar{\Y}_i^s\|_F^2,\nonumber
\end{eqnarray}
\vspace{-5pt}

\noindent in which $\bar{\X}_i^{s}\in\mathcal{R}^{d\times C^s}$ is horizontally stacked $C^s$ copies of $\x_i^s$ and $\bar{\Y}_i^s\in\mathcal{R}^{C^s\times C^s}$ is vertically stacked $C^s$ copies of $\y_i^s$. Hence, in our implementation, given the $i$-th training instance, we duplicate $\x_i^s$ and $\y_i^s$ to $C^s$
copies, and apply the aggregated feature masks $\M^s$ over all $C^s$ categories on the duplicated visual features $\bar{\X}_i^{s}$. Then, the decision value matrix of the $i$-th training instance $({\bar{\X}_i^{s'}}\circ \M^{s'})\W\A^s$ can be easily obtained.

\noindent\textbf{Specificity: }On the other hand, we expect that $\bar{\W}_c^s$ can better fit the classification task corresponding to the $c$-th category. For the $i$-th training instance, we use $\J^i=({\bar{\X}_i^{s'}}\circ \M^{s'})\W\A^s$ to denote its decision value matrix, in which $J^i_{c_1,c_2}$ is the decision value of $c_2$-th category obtained by using $\bar{\W}_{c_1}^s$.
In the following, we only consider the decision values of its ground-truth category $c(i)$, \ie, the $c(i)$-th column of $\J^i$. In this case, similar as in (\ref{eqn:W_adaptation}), the decision value of $c(i)$-th seen category obtained by using $\bar{\W}_{\tilde{c}}^s$ is 

\vspace{-15pt}
\begin{eqnarray} \label{eqn:J_ij}
J^i_{\tilde{c},c(i)}= (\x_i^{s'}\circ \m_{\tilde{c}}^{s'})\W\a^s_{c(i)} = \x_i^{s'}\bar{\W}_{\tilde{c}}^s \a_{c(i)}^s.
\end{eqnarray}

\noindent Then, we expect $J^i_{c(i),c(i)}$ obtained by $\bar{\W}_{c(i)}^s$ should be larger than $J^i_{\tilde{c},c(i)}$ obtained by $\bar{\W}_{\tilde{c}}^s$ for $\tilde{c}\neq c(i)$. With this aim, we employ the hinge loss $R^i = \sum_{\tilde{c}\neq c(i)} \max(0, J^i_{\tilde{c},c(i)}-J^i_{c(i),c(i)}+\rho)$ to push $J^i_{c(i),c(i)}$ to be larger than $J^i_{\tilde{c},c(i)}$ by margin $\rho$ for $\tilde{c}\neq c(i)$. In our experiments, we empirically set $\rho$ as $0.5$ considering that $\J^i$ is regressed to binary label matrix.

By taking both generalizability and specificity into consideration, the loss function of our deep adaptive embedding model is designed as

\vspace{-15pt}
\begin{eqnarray} \label{eqn:deep_loss}
\min_{g(\cdot), \W}\sum_{i=1}^{n^s} (\|\J^i-\bar{\Y}_i^s\|_F^2 + R^i).
\end{eqnarray}
Based on (\ref{eqn:deep_loss}), we aim to have implicit $\bar{\W}^s_c$'s which can generalize to other categories by minimizing $\|\J^i-\bar{\Y}_i^s\|_F^2$ and simultaneously better fit the classification task corresponding to its own category by minimizing $R^i$. 
Note that semantic representations of unseen categories and unlabeled test instances are not utilized in the training stage.

Our deep adaptive embedding model is illustrated in Fig.~\ref{fig:DAEZSL}, from which we can see that
$g(\cdot)$ is modeled by multi-layer perceptrons (MLP) and $\W$
is modeled by fully connected (fc) layer. Specifically, during the training process, we input the training images and the semantic representations of all $C^s$ seen categories, \ie, ${\A^s}'\in\mathcal{R}^{C^s\times a}$. In the top flow, $d$-dim feature of the $i$-th training image is duplicated to $C^s$ copies, \ie, ${\bar{\X}_i^{s'}}\in\mathcal{R}^{C^s\times d}$. In the bottom flow,  ${\A^s}'$ passes through MLP and generates the feature masks $\M^{s'}\in\mathcal{R}^{C^s\times d}$ for $C^s$ categories. After applying the generated feature masks on the duplicated visual features via elementwise product, the masked features ${\bar{\X}_i^{s'}}\circ\M^{s'}$ are fed into fc layer (\ie, the common visual-semantic mapping matrix $\W$), and output $({\bar{\X}_i^{s'}}\circ \M^{s'})\W$.
Finally, we perform dot product on $({\bar{\X}_i^{s'}}\circ \M^{s'})\W$ and $\A^s$, leading to the decision value matrix $\J^i$, based on which we employ the loss function in (\ref{eqn:deep_loss}). Note that we train an end-to-end system to minimize the loss function in (\ref{eqn:deep_loss}), during which the parameters of CNN, MLP, and fc layer in Fig.~\ref{fig:DAEZSL} are jointly optimized.  More details about the network architecture and training process can be found in Section~\ref{sec:exp_large}.

In the prediction stage, given an unseen category with semantic representation $\a^t_{\tilde{c}}$, we can easily generate its feature mask $\m^t_{\tilde{c}}$ via the projection function $g(\cdot)$, which implies an adaptive visual-semantic mapping $\bar{\W}^t_{\tilde{c}}$.
Intuitively, if $\a^t_{\tilde{c}}$ is similar to $\a^s_c$ of the $c$-th seen category, their generated feature masks $\m^t_{\tilde{c}}$ and $\m^s_c$ should be similar. Furthermore, their  visual-semantic mapping matrices $\bar{\W}^t_{\tilde{c}}$ and $\bar{\W}^s_c$ should be close to each other. Therefore, the visual-semantic mapping of a given unseen category is expected to be in correlation with those of similar seen categories, analogous to learning $\W^c$ based on the similarity matrix $\S^c$ in (\ref{eqn:ESZSL_category}).

We further elaborate on the prediction procedure based on Fig.~\ref{fig:DAEZSL}. In particular, given the $i$-th test image $\x_i^t$ and the set of unseen categories with size $C^t$, we use this test image and the semantic representations of all $C^t$ unseen categories ${\A^t}'\in\mathcal{R}^{C^t\times a}$ as input. The test image passes through the top flow and generates $C^t$ duplicated copies of visual features, \ie, $\bar{\X}_i^{t'} \in \mathcal{R}^{C^t\times d}$, while ${\A^t}'$ passes through the bottom flow and generates the feature masks $\M^{t'}\in\mathcal{R}^{C^t\times d}$ for all unseen categories. After performing $({\bar{\X}_i^{t'}}\circ \M^{t'})\W\A^t$, we can obtain a decision value matrix $\J^i\in\mathcal{R}^{C^t\times C^t}$ for the $i$-th test image. Finally, we get the diagonal of $\J^i$ as decision value vector and classify this test image as the category corresponding to the highest decision value. Note that only the diagonal of $\J^i$ is used for prediction because we assume that for the $c$-th unseen category, the decision value $J^i_{c,c}=\x_i^{t'}\bar{\W}_c^t\a_c^t$ (see (\ref{eqn:J_ij})) obtained based on $\bar{\W}_c^t$ can best fit the classification task for this category.

\noindent\textbf{Relation to ESZSL: }When fixing the feature masks for all categories, \ie, $\M^s$, as all-one matrix without learning MLP in Fig.~\ref{fig:DAEZSL}, our DAEZSL model approximately reduces to ESZSL, in which the visual-semantic mappings of all categories are the same. 

\noindent\textbf{Relation to AEZSL: }Our DAEZSL model shares the similar idea with our AEZSL method, that is, visual-semantic mapping should be category-specific and related to the semantic representation of each category. Besides, we design the training loss in (\ref{eqn:deep_loss}) considering the generalizability and specificity of visual-semantic mappings, in analogy to learning category-specific $\W^c$'s with co-regularizer in (\ref{eqn:ESZSL_category}). Moreover, in our DAEZSL model, the visual-semantic mapping of a given unseen category is expected to be close to those of seen categories with similar semantic representations, resembling the first regularizer based on the similarity matrix $\S^c$ in (\ref{eqn:ESZSL_category}).

\section{Experiments} \label{sec:exp}
In this section, we conduct experiments for image classification on three small-scale datasets (\eg, CUB, SUN, and Dogs) and one large-scale dataset (\eg, ImageNet). On the small-scale datasets, we compare our AEZSL and AEZSL\_LR methods with their special cases as well as standard/semi-supervised ZSL baseline methods. Note that our DAEZSL method is designed for large-scale ZSL and the small-scale training set can easily cause overfitting, especially for small number of seen categories, so we omit the results of DAEZSL on the small-scale datasets.
On the large-scale dataset, due to the inefficiency of AEZSL for a large number of unseen categories, we only evaluate our DAEZSL method, which is specifically designed for large-scale ZSL, and compare with recently reported state-of-the-art results.

\subsection{Zero-Shot Learning on Small-Scale Datasets} \label{sec:exp_standard}

\noindent\textbf{Experimental Settings:} We conduct experiments on the following three popular benchmark datasets which are commonly used for zero-shot learning tasks:
\begin{itemize}
\item  CUB~\cite{WahCUB_200_2011}: Caltech-UCSD Bird (CUB) has in total $11,788$ images distributed in 200 bird categories. Following the ZSL setting in \cite{akata2013label}, we use the standard train-test split with $150$ (\resp, $50$) categories as seen (\resp, unseen) categories. The CUB dataset contains a $312$-dim binary human specified attribute vector for each image, so we average the attribute vectors of the images within each category and use it as the semantic representation of that category.
\item  SUN~\cite{xiao2010sun}: Scene UNderstanding (SUN) dataset has 20 images in each scene category. Following the ZSL setting in \cite{jayaraman2014zero}, we use the provided list of $10$ categories as unseen categories and the rest of 707 categories as seen categories. Similar to CUB, the averaged $102$-dim attribute vector is used as the semantic representation for each category.
\item Dogs~\cite{KhoslaYaoJayadevaprakashFeiFei_FGVC2011}: ZSL was firstly performed on the Stanford Dogs dataset in \cite{akata2015evaluation}, which uses $19,501$ images from $113$ breeds of dogs. We follow the provided train-test split in \cite{akata2015evaluation}, \ie, $85$ (\resp, $28$) categories as seen (\resp, unseen) categories. Since there is no manually annotated attribute for the Dogs dataset, we combine two types of output embeddings learned from online corpus (\ie, $3,850$-dim Bag-of-Words embedding and $163$-dim WordNet-derived similarity embedding) as the semantic representation for each category, which has demonstrated superior performance in~\cite{akata2015evaluation}. 
\end{itemize}

\setlength{\textfloatsep}{5pt}
\begin{table}[t]
\caption{Accuracies (\%) of different baseline methods and our methods on three benchmark datasets. The best results are highlighted in boldface.}
\setlength{\tabcolsep}{5pt}
\label{tab:exp_ZSL}
\centering
%\small
\begin{tabular}{|c|c|c|c|c|}
\hline
Dataset & CUB & SUN & Dogs & Avg \\
\hline
ESZSL~\cite{romera2015embarrassingly} & 49.74 & 82.50 & 40.27 & 57.50 \\
LatEm~\cite{xian2016latent} & 45.50	& 83.50	& 37.41	& 55.47\\
Zhang and Saligrama~\cite{zhang2016zerose} & 46.11 & 83.83 & 43.42 & 57.79\\
Bucher~\etal~\cite{bucher2016improving} & 43.29	& 84.41	& 36.18	& 54.63\\
AMP~\cite{fu2015zero} & 43.12 &	82.50 & 38.02 &	54.55 \\
COSTA~\cite{mensink2014costa} & 44.19 & 76.00 & 33.98 & 51.39\\
SSE~\cite{zhang2015zero} & 40.64 & 82.50 & 38.63 & 53.92\\
SJE~\cite{akata2015evaluation} & 51.70 & 84.00 & 37.11 & 57.60\\
DAP/IAP~\cite{lampert2014attribute} & 41.40 & 72.00 & 37.11 & 50.17\\
Changpinyo~\etal~\cite{changpinyo2016synthesized} & 54.50 & 83.50 & 42.01 & 60.50\\
ConSE~\cite{norouzi2013zero} & 37.33 & 73.00 & 23.49 & 44.61 \\
RKT~\cite{wang2016relational} & 53.73 & 81.00 & 42.11 & 58.95 \\
EXEM~\cite{changpinyo2016predicting} & 58.54 & 86.00 & 42.90 & 62.48\\
\hline
Li~\etal~\cite{li2015semi} & 43.94 & 87.50	& 43.50 & 58.31\\
Kodirov~\etal~\cite{kodirov2015unsupervised} & 57.42 & 86.00 & 48.59 & 64.00\\
Zhang and Saligrama~\cite{zhang2016zero} & 56.90 & 89.50 & 48.53 & 64.98 \\
Shojaee and Baghshah~\cite{shojaee2016semi} & 58.80 & 86.16 & 47.49 & 64.15\\
Li and Guo~\cite{li2015max} & 57.14 & 88.50 & 47.84 & 64.49 \\
SMS~\cite{guo2016transductive} & 57.14 & 83.50 & 47.95 & 62.86 \\
Xu~\etal~\cite{xu2017transductive} & 55.74 & 85.50 & 43.09 & 61.44\\
\hline
AEZSL\_sim & 57.55	& 87.50	& 43.98	& 63.01\\
AEZSL & 59.73 & 88.00 & 44.62 & 64.12\\
\hline
AEZSL\_LR (one-step) & 60.08 & 89.00 & 48.90 & 65.99\\
AEZSL\_LR & \textbf{64.44} & \textbf{92.50} & \textbf{50.62} & \textbf{69.18}\\
\hline
\end{tabular}
\end{table}

For CUB and SUN datasets, we extract the $4,096$-dim output of the $6$-th layer of the pretrained VGG~\cite{simonyan2014very} as the visual feature for each image, following myriad of previous ZSL works such as \cite{zhang2016zerose,zhang2016zero,zhang2015zero,bucher2016improving}. For the Dogs dataset, we use the $1,024$-dim output of the top layer of the pretrained
GoogleNet \cite{szegedy2015going} following~\cite{xian2016latent, akata2015evaluation}.

\noindent\textbf{Baselines:} We compare our AEZSL and AEZSL\_LR methods with two sets of ZSL baselines: standard ZSL methods and semi-supervised/transductive ZSL methods. For the first set, we include the following methods, \cite{xian2016latent,zhang2016zerose,bucher2016improving,fu2015zero,mensink2014costa,zhang2015zero,
akata2015evaluation,lampert2014attribute,changpinyo2016synthesized,norouzi2013zero,wang2016relational,changpinyo2016predicting}, which do not utilize unlabeled test instances in the training stage. For the second set, we compare with the following transductive or semi-supervised ZSL methods \cite{li2015semi, kodirov2015unsupervised, zhang2016zero, shojaee2016semi,li2015max,guo2016transductive,xu2017transductive}, which utilize the unlabeled test instances and semantic representations of unseen categories in the training phase. Note that our AEZSL belongs to standard ZSL methods while AEZSL\_LR belongs to semi-supervised ZSL methods.

Besides two sets of baselines mentioned above, we consider a special case of our AEZSL method, \ie, without using the co-regularizer $\sum_{c<\tilde{c}}\|\W^c-\W^{\tilde{c}}\|_F^2$ ($\lambda_3=0$), which is referred to as AEZSL\_sim in Table~\ref{tab:exp_ZSL}. We also report the results of one special cases of our AEZSL\_LR method, namely, AEZSL\_LR (one-step), which is a non-progressive approach by treating the entire test set as the unconfident set $\mathcal{U}$ and using the method in (\ref{eqn:EZSL_refine}) without the regularizer $\|{\X^l}'\P-\Y^l\|_F^2$.

We use multi-class accuracy for performance evaluation. For the baselines, if the experimental settings (\ie, train-test split, visual feature, semantic representation, evaluation metric, \etc) mentioned in their papers are exactly the same as ours, we directly copy their reported results. Otherwise, we run their methods using our experimental setting for fair comparison.

\noindent\textbf{Parameters:} Our AEZSL method has three trade-off parameters $\lambda_1$, $\lambda_2$, and $\lambda_3$ in (\ref{eqn:ESZSL_category}). Besides, our label refinement strategy has three additional trade-off parameters $\gamma_1$, $\gamma_2$, and $\gamma_3$ (see (\ref{eqn:EZSL_refine})). We use cross-validation strategy to determine the above trade-off parameters.
Specifically, following~\cite{shojaee2016semi},  we choose the first $C^c$ categories based on the default category indices from $C^s$ seen categories as validation categories, in which $C^c$ satisfies $\frac{C^c}{C^s}=\frac{C^t}{C^s+C^t}$. In our experiments, we first learn models based on the seen categories excluding the validation categories with $\lambda_1$, $\lambda_2$, and $\lambda_3$ set within the range $[10^{-3},10^{-2},\ldots,10^{3}]$, and then test the learnt models on the validation categories. After determining the optimal trade-off parameters through the grid search, we learn the final model based on all seen categories.  
Note that we also have a hyper parameter $k$, which is the number of selected confident test instances in each iteration. We empirically fix $k$ as $100$ for CUB and Dogs datasets, and $10$ for the SUN dataset given that there are only $200$ test instances in the SUN dataset. 

\begin{figure*}[t]
        \centering
        \includegraphics[width=0.32\textwidth]{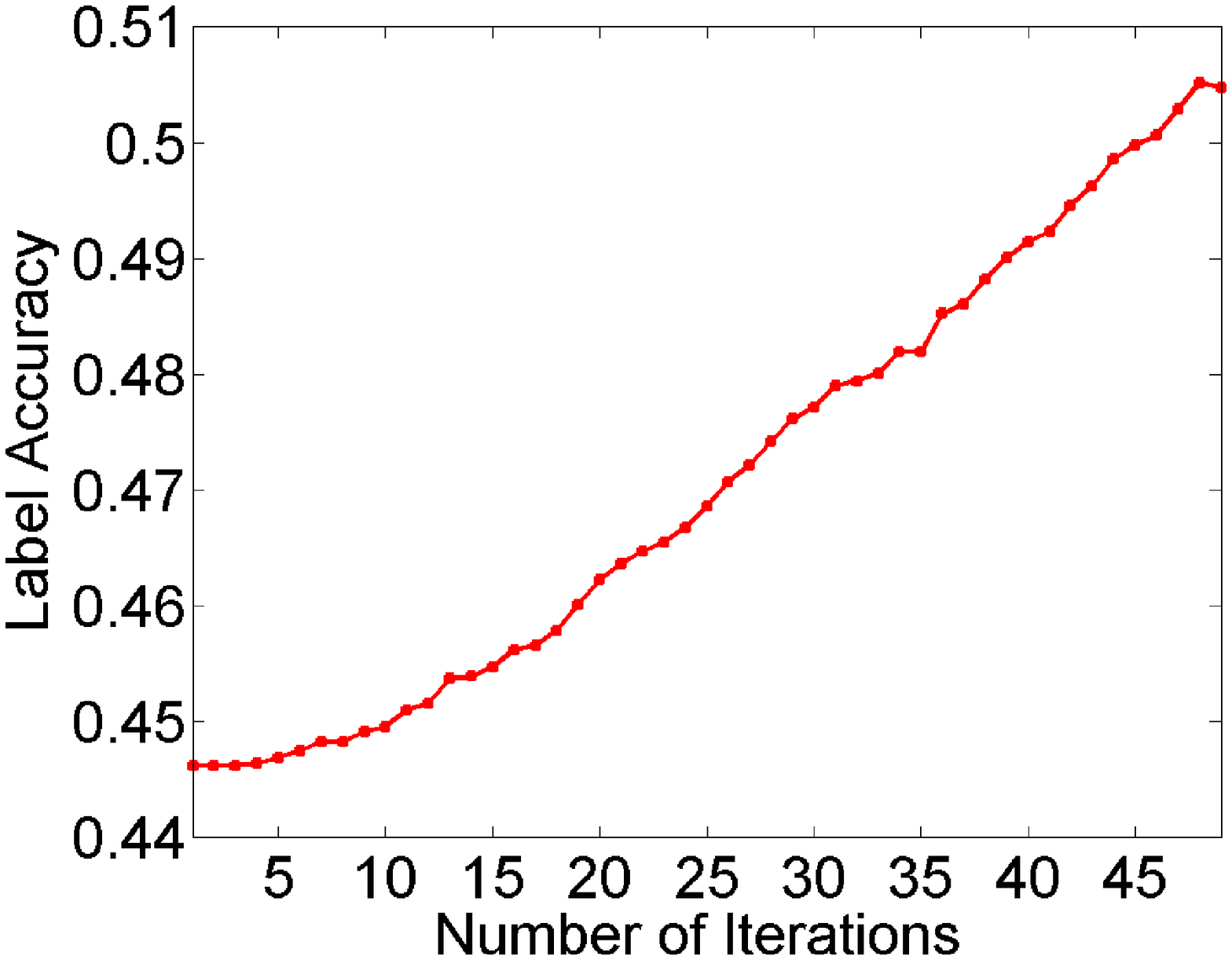}
		\includegraphics[width=0.32\textwidth]{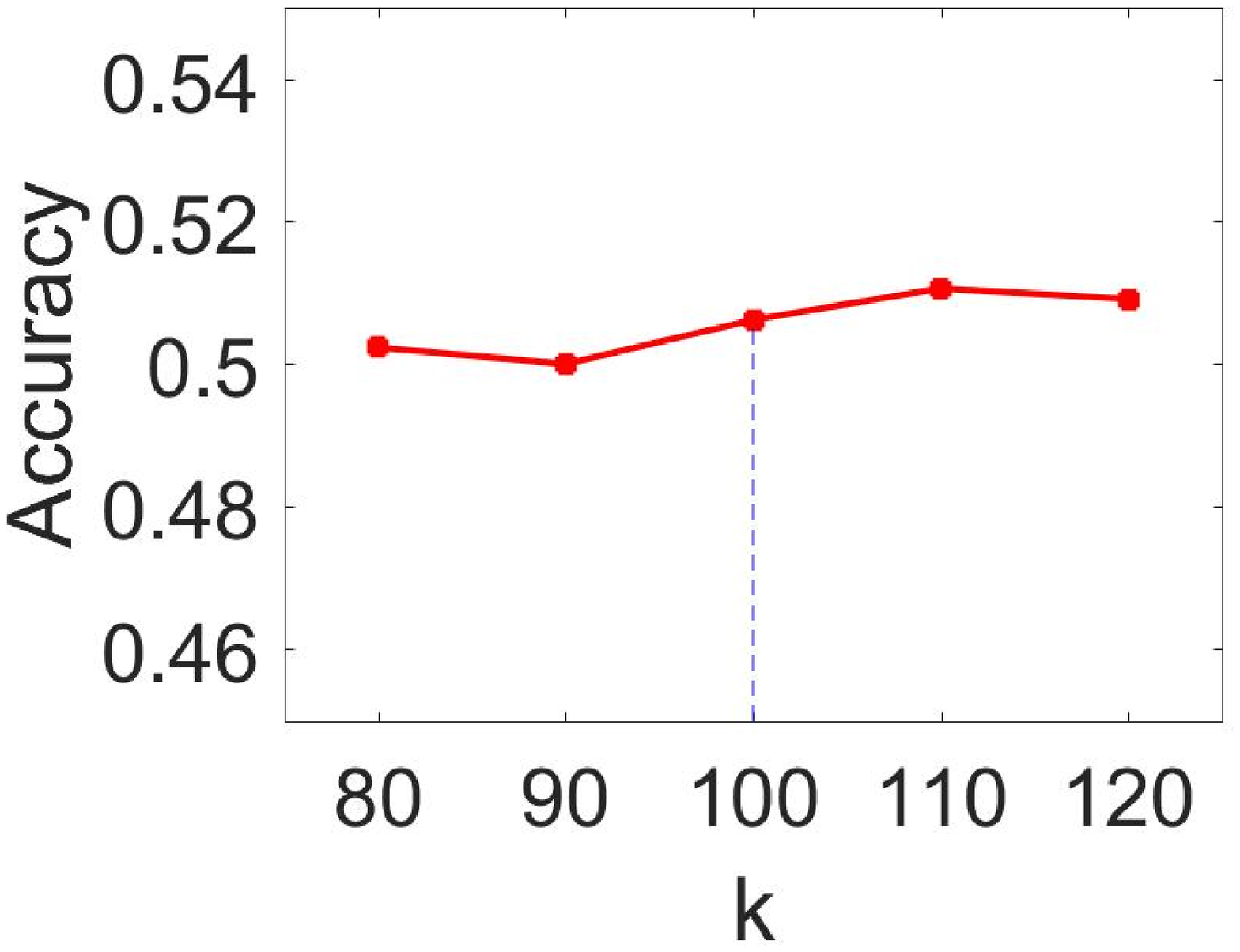}
		\includegraphics[width=0.32\textwidth]{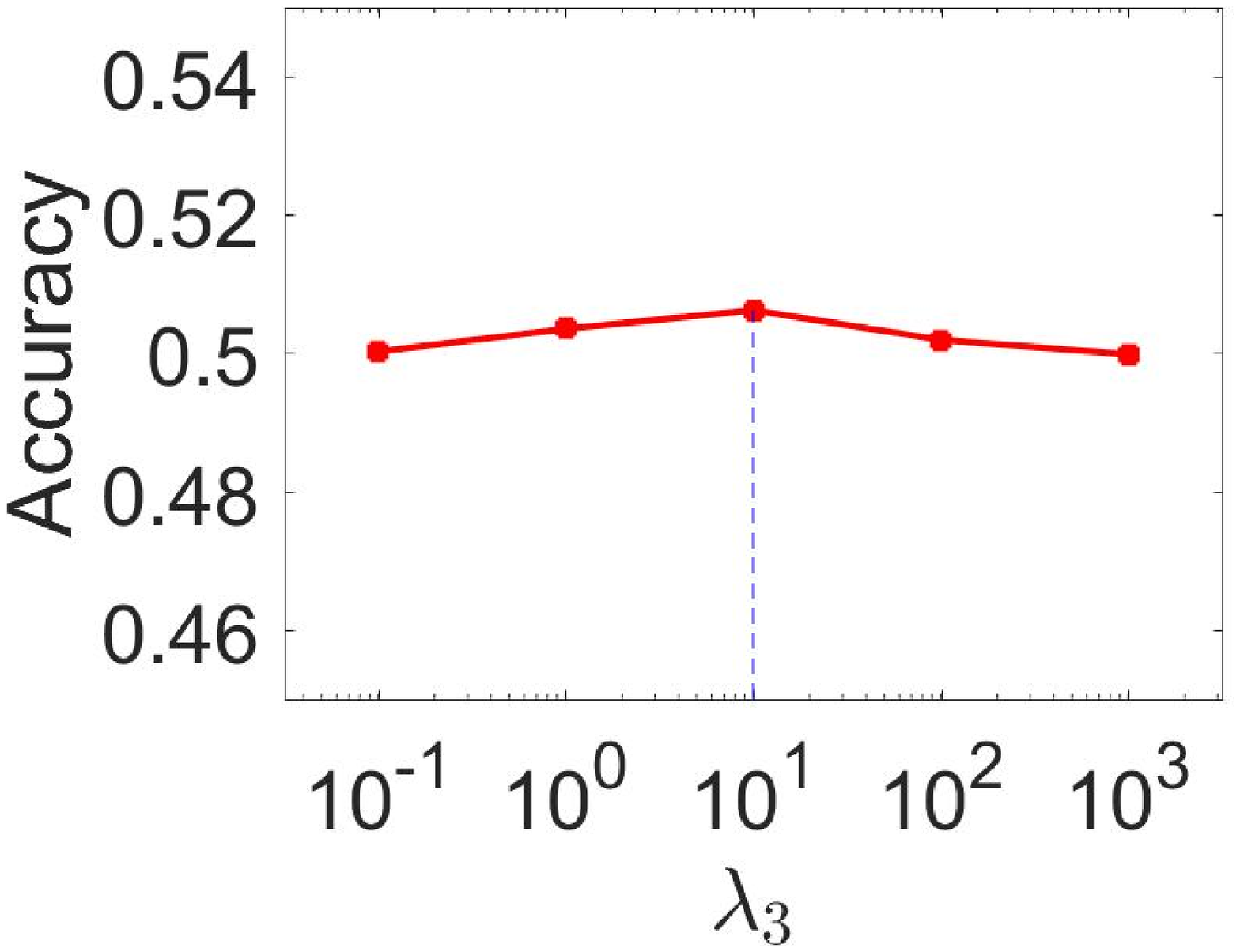}     
        
        \caption{The left subfigure shows the accuracy variation of predicted test labels \wrt the number of iterations in label refinement on the Dogs dataset. The middle (\resp, right) subfigure shows the performance variance of our AEZSL\_LR method \wrt parameter $k$ (\resp, $\lambda_3$) on the Dogs dataset, in which the dash line indicates the parameter we use in Table~\ref{tab:exp_ZSL}.}
        \label{fig:progressive_refinement}
        \vspace{-10pt}
\end{figure*}

Based on our experimental observation, our methods are relatively robust when the parameters are set within certain range. By taking $\lambda_3$ and $k$ as examples, we vary $\lambda_3$ (\resp, $k$) in the range of $[10^{-1},\ldots, 10^3]$ (\resp, $[80, 90, \ldots, 120]$) and evaluate our AEZSL\_LR method on the Dogs dataset. The performance variance is illustrated in the middle and right subfigure in Fig.~\ref{fig:progressive_refinement}, from which we can see that our method is insensitive to $\lambda_3$ and $k$ within certain range. We have similar observations for the other parameters on the other datasets.

\noindent\textbf{Experimental Results:} The experimental results for our AEZSL and AEZSL\_LR methods as well as all the baselines are reported in Table~\ref{tab:exp_ZSL}. 
It can be seen that ESZSL achieves competitive results compared with other standard ZSL baselines, which demonstrates the effectiveness of ESZSL despite its simplicity. By comparing AEZSL\_sim with AEZSL, we observe that AEZSL achieves better results after employing the co-regularizer, so it is beneficial to encourage the mappings from different categories to share some common parts. We also observe that our AEZSL method not only outperforms ESZSL and AEZSL\_sim, but also outperforms the standard ZSL baselines~\cite{xian2016latent,zhang2016zerose,bucher2016improving,fu2015zero,mensink2014costa,zhang2015zero,
akata2015evaluation,lampert2014attribute,changpinyo2016synthesized,
norouzi2013zero,wang2016relational,changpinyo2016predicting}, which indicates the advantage of learning a visual-semantic mapping for each unseen category.  

From Table~\ref{tab:exp_ZSL}, we also observe that the transductive or semi-supervised baselines~\cite{li2015semi, kodirov2015unsupervised, zhang2016zero, shojaee2016semi,li2015max,guo2016transductive,xu2017transductive} generally achieve better results than those standard ZSL baselines, indicating that it is helpful to utilize the unlabeled test instances in the training stage. Another observation is that our AEZSL\_LR method outperforms AEZSL, which demonstrates the effectiveness of the progressive label refinement. Moreover, AEZSL\_LR also performs better than AEZSL\_LR (one-step). This is because the selected confident set with more accurate predicted labels can provide weak supervision.
Finally, our AEZSL\_LR method outperforms all the baselines and achieves the state-of-the-art results on three datasets, which again shows the advantage of adapting the visual-semantic mapping to each unseen category. 

\setlength{\textfloatsep}{0pt}
\begin{figure}[t]
        \centering
        \begin{subfigure}[b]{0.49\textwidth}
                \includegraphics[width=0.48\textwidth]{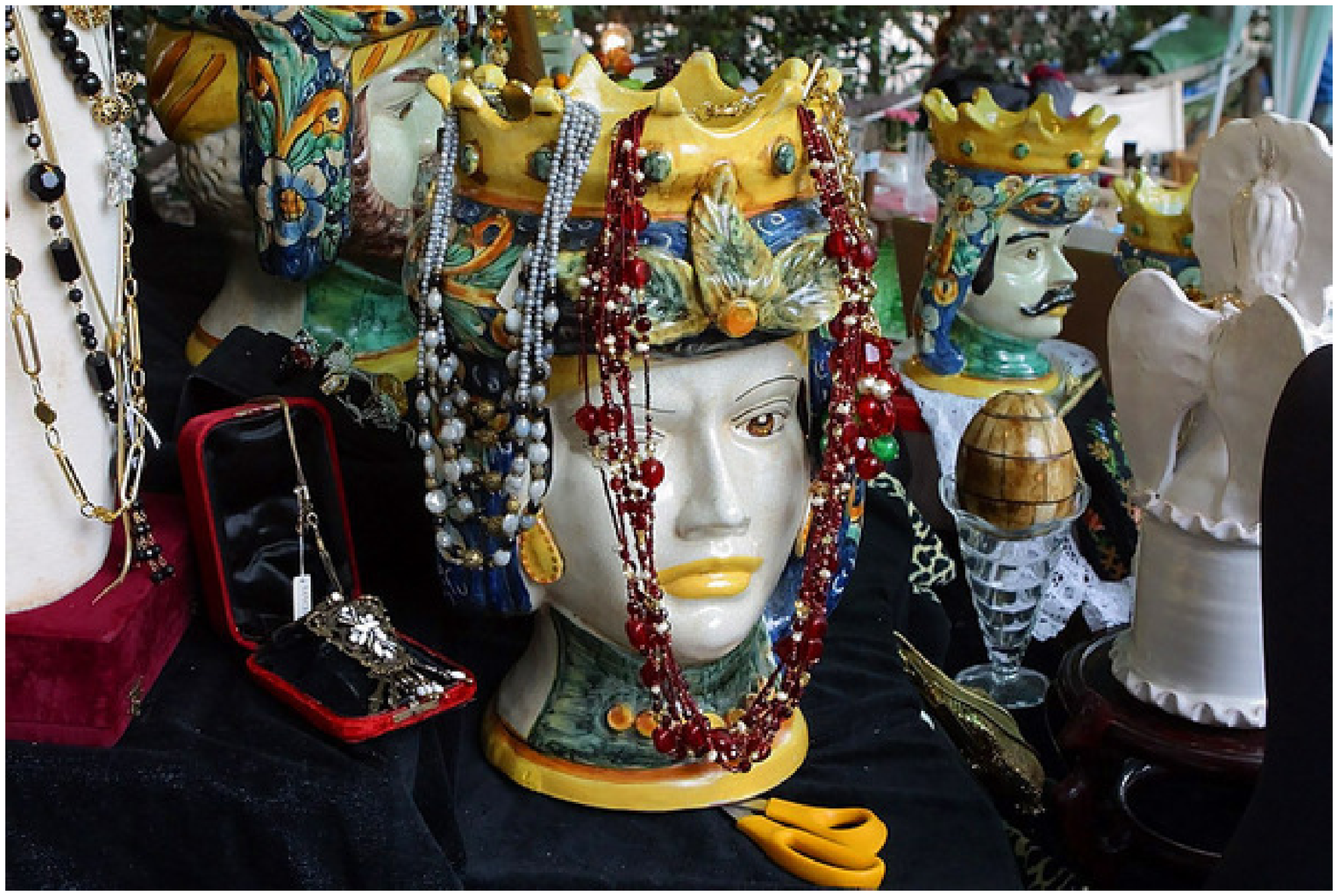}
                \includegraphics[width=0.50\textwidth]{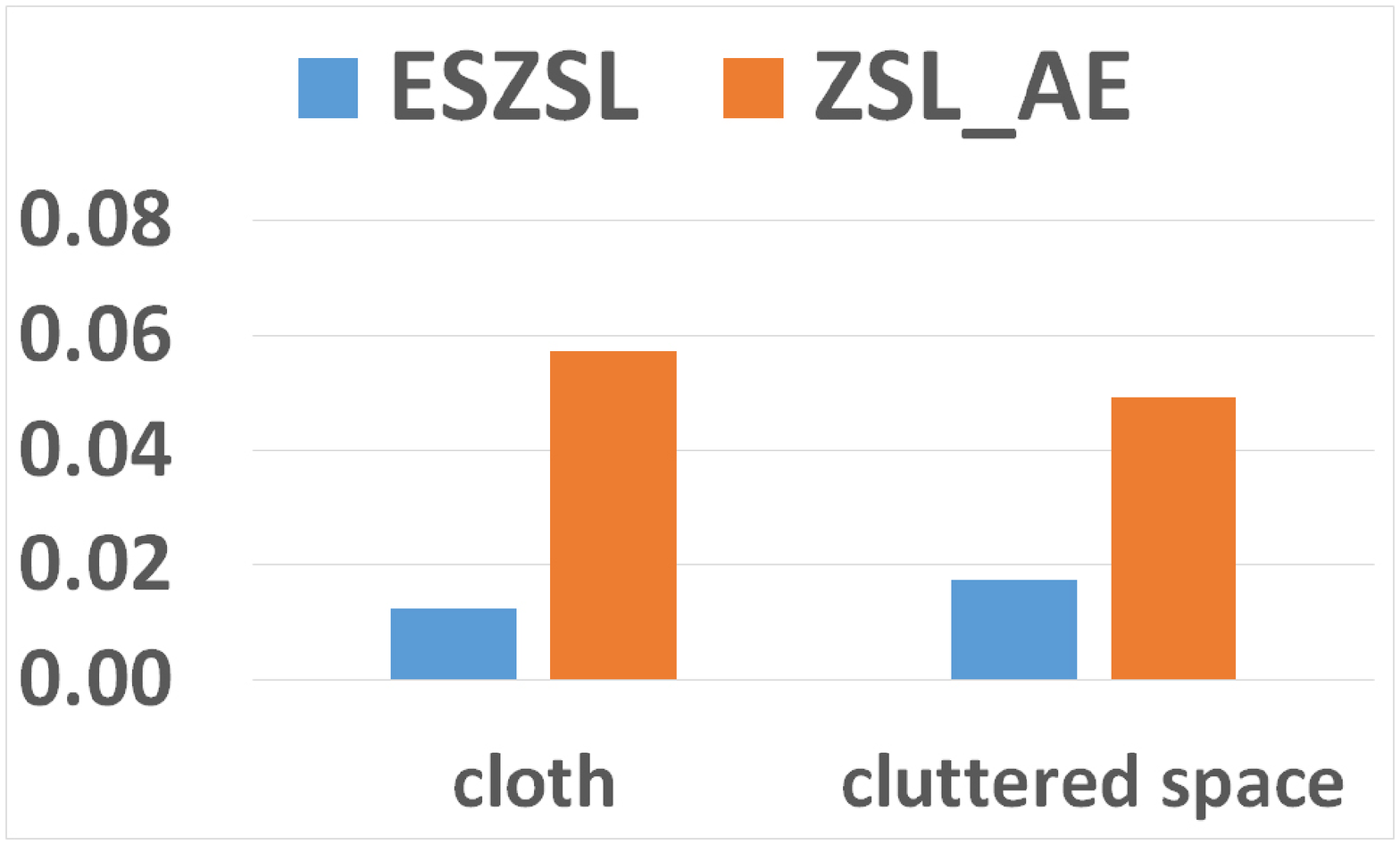}                
        \end{subfigure}
        \begin{subfigure}[b]{0.49\textwidth}
                \includegraphics[width=0.48\textwidth]{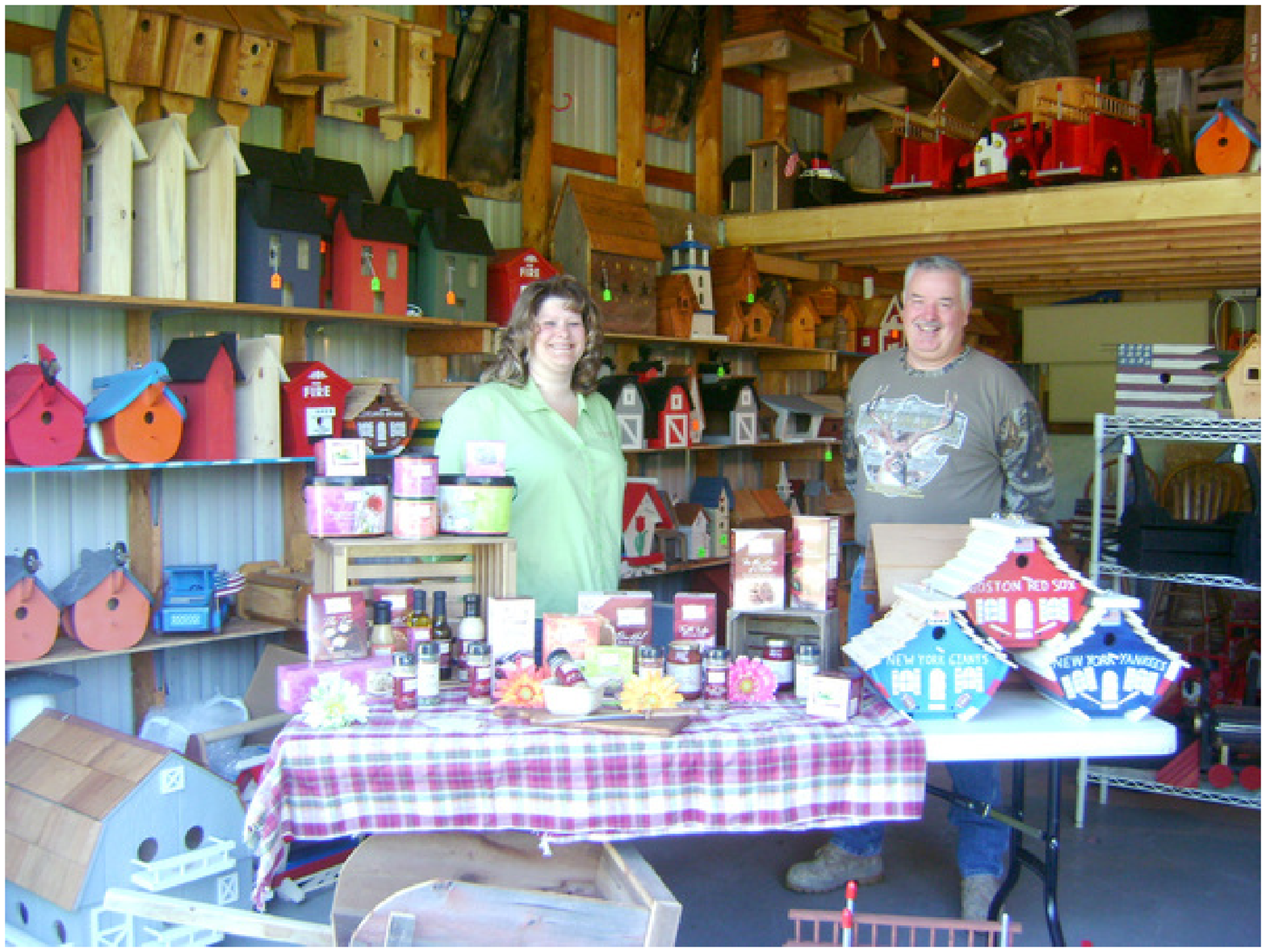}       
                \includegraphics[width=0.50\textwidth]{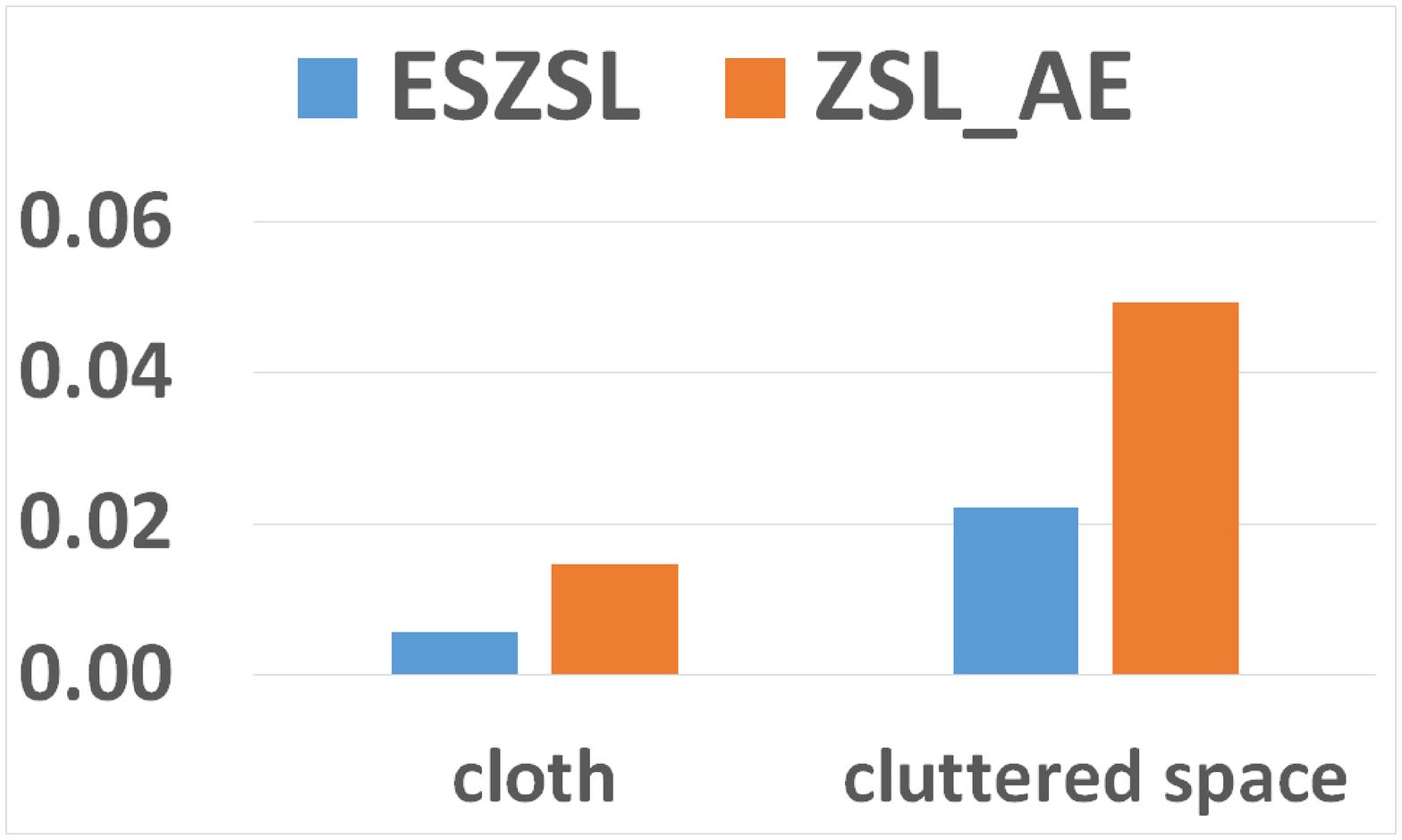}
        \end{subfigure}
        \caption{Illustration of two instance images from the category ``flea market" and their corresponding mapped values for the attributes ``cloth" and ``cluttered space" obtained by using ESZSL and our AEZSL method.}
        \label{fig:showcases}
\end{figure}
\setlength{\textfloatsep}{5pt}
\begin{figure}[t]
        \centering
        \begin{subfigure}[b]{0.2385\textwidth}
                \includegraphics[width=0.95\textwidth]{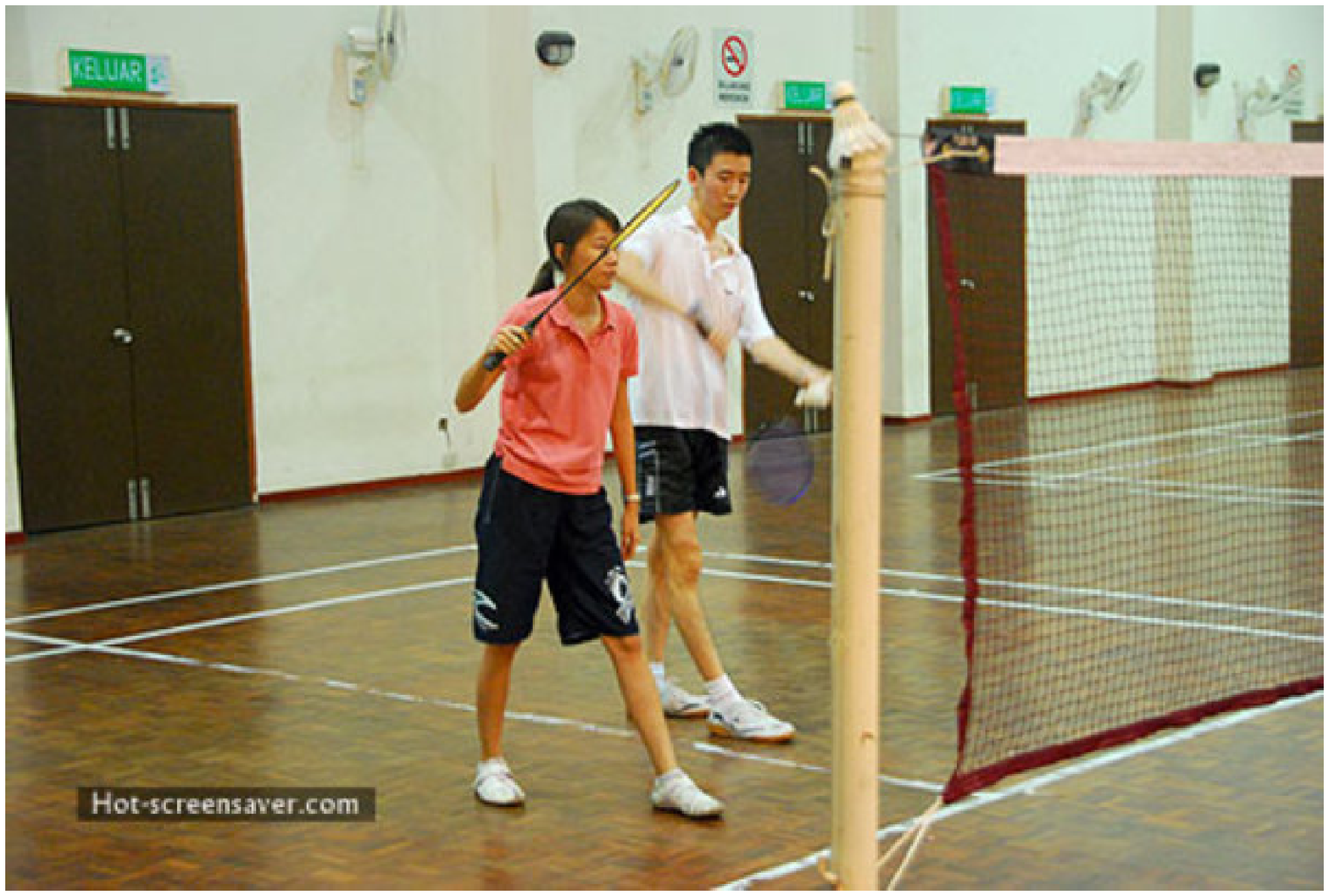}
                \caption{badminton court (indoor)}
        \end{subfigure}
        \begin{subfigure}[b]{0.2415\textwidth}
                \includegraphics[width=0.95\textwidth]{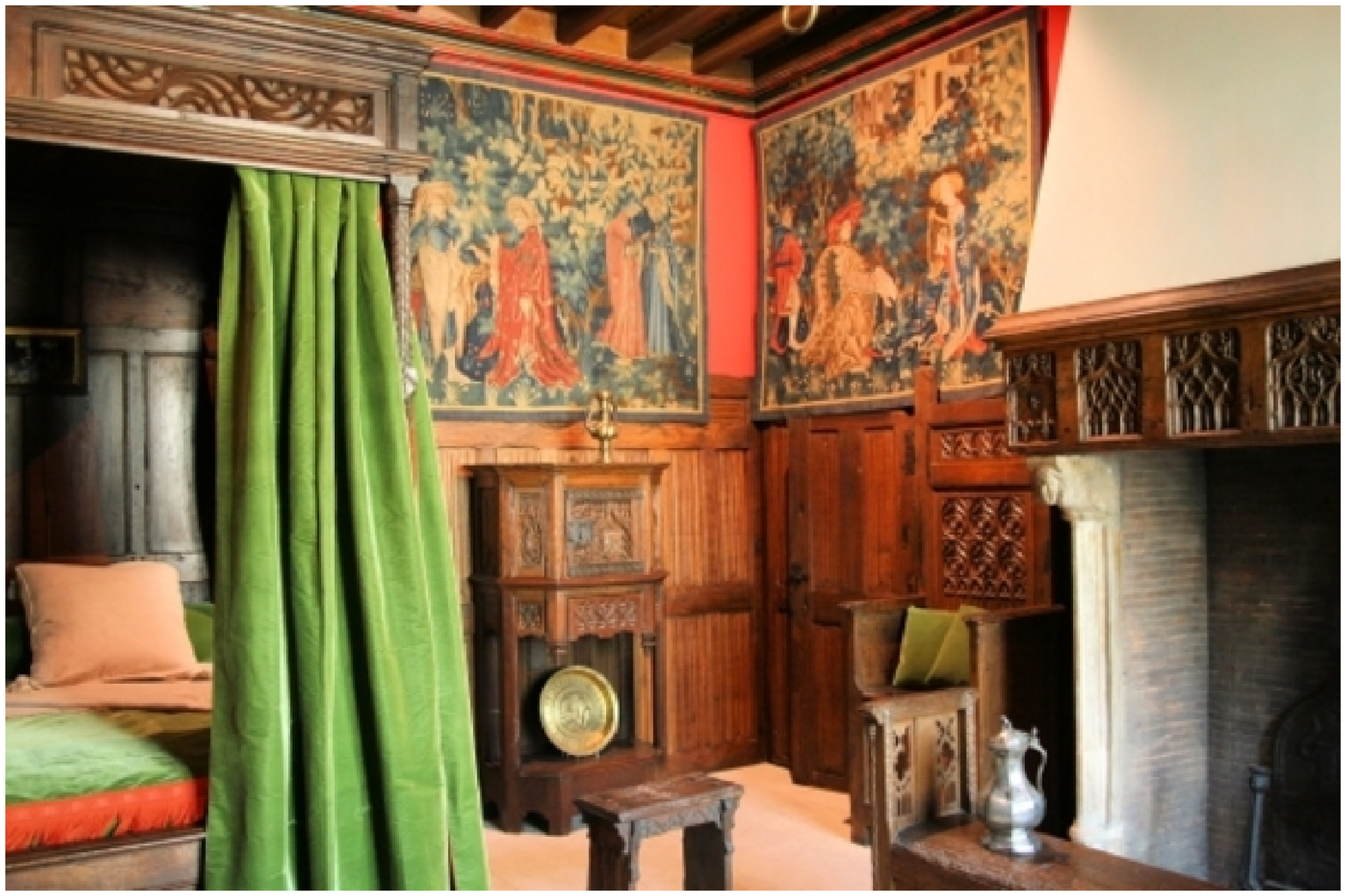}
                \caption{bedchamber}
        \end{subfigure}
        \begin{subfigure}[b]{0.2234\textwidth}
                \includegraphics[width=0.95\textwidth]{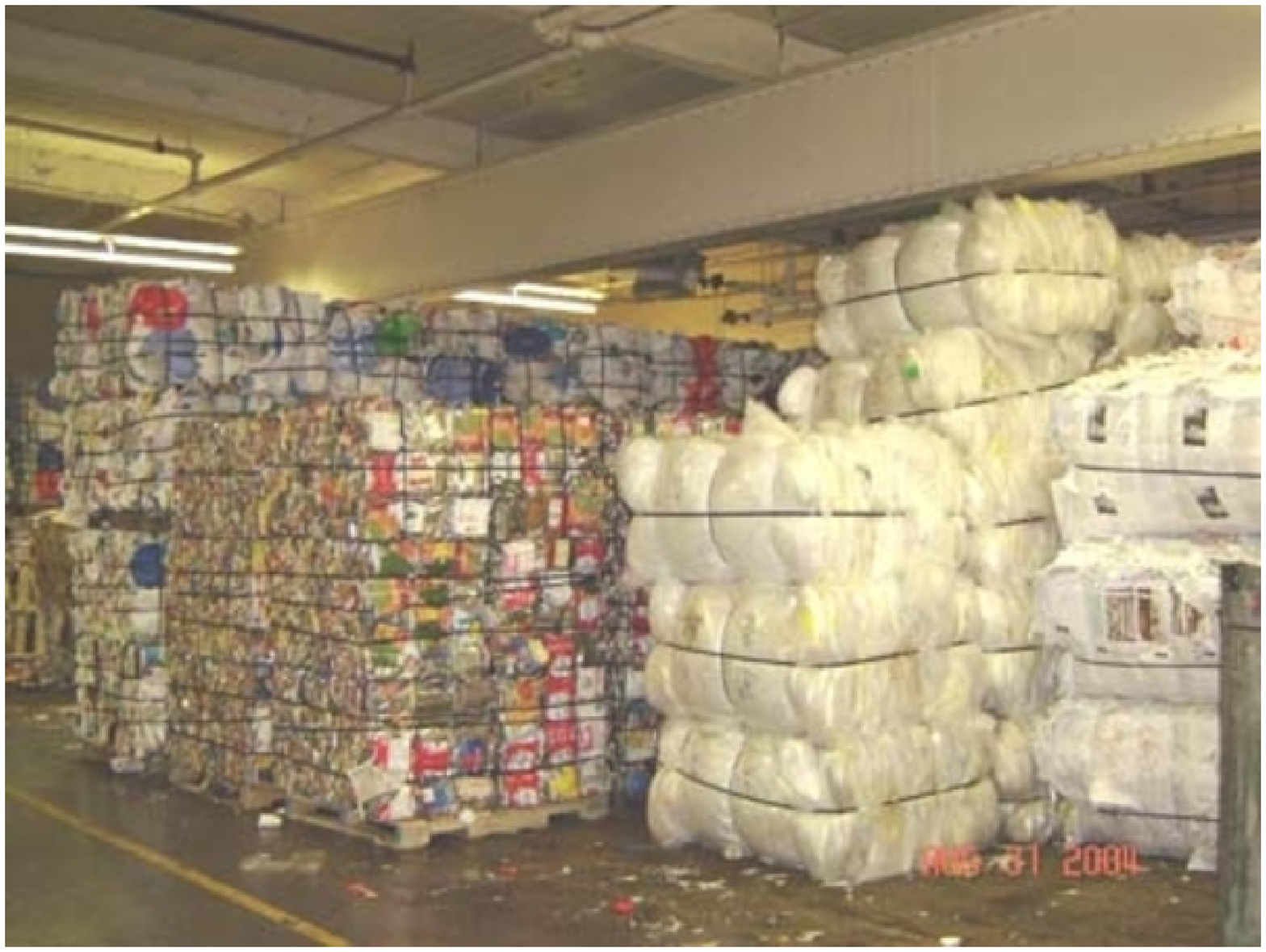}               
                \caption{recycling plant}
        \end{subfigure}
        \begin{subfigure}[b]{0.2566\textwidth}
                \includegraphics[width=0.95\textwidth]{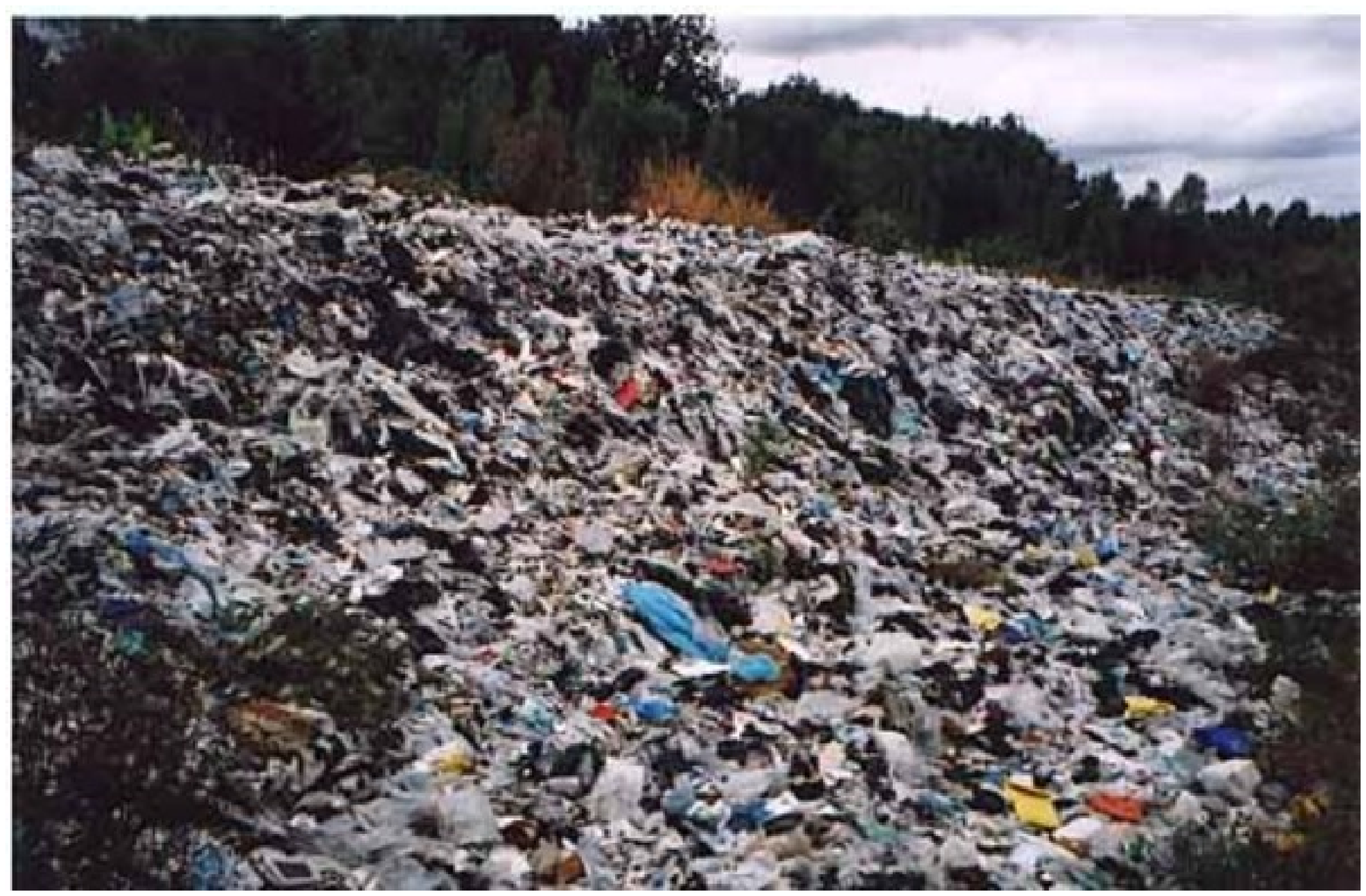}
                \caption{landfill}
        \end{subfigure}
        \caption{The first (\resp, second) row contains the instance images from different categories with the attribute ``cloth" (\resp, ``cluttered space").}
        \label{fig:attributes}
\end{figure}

\setlength{\textfloatsep}{5pt}
\begin{figure*}[t]
        \centering
        \begin{subfigure}[b]{0.2985\textwidth}
                \includegraphics[width=1\textwidth]{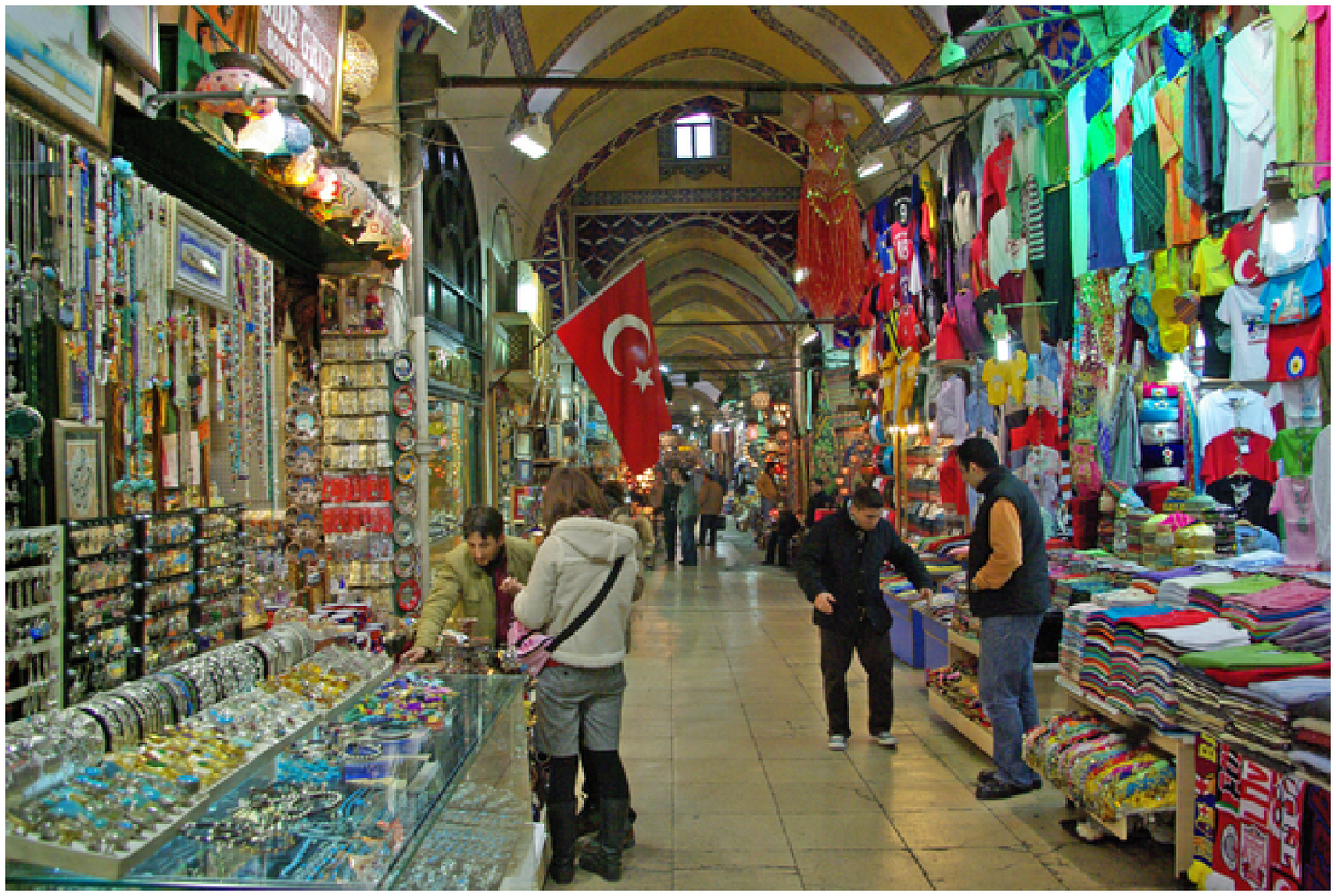}
				\caption{bazzar (indoor)}
        \end{subfigure}
        \begin{subfigure}[b]{0.1496\textwidth}
                \includegraphics[width=1\textwidth]{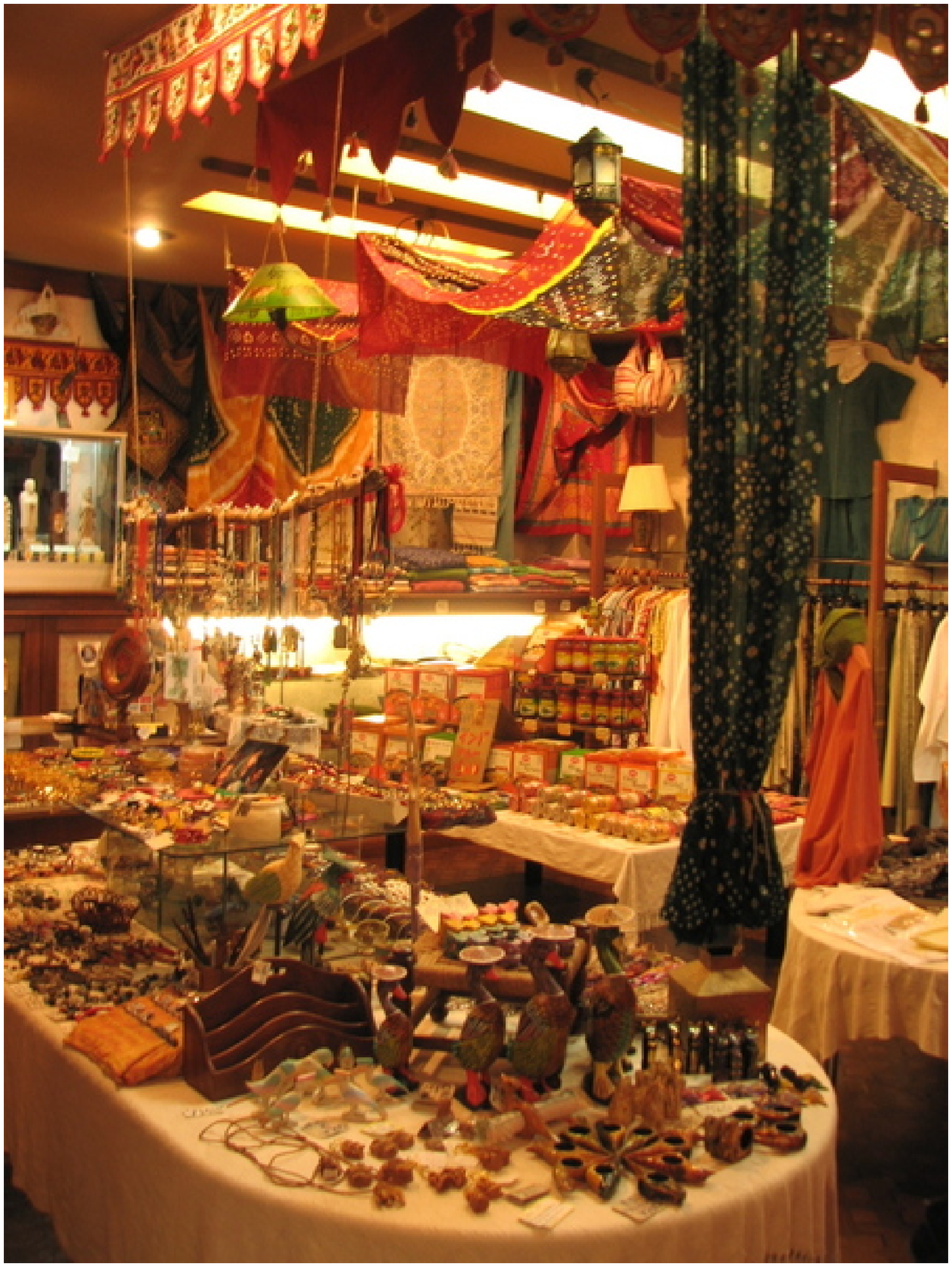}
				\caption{thrift shop}
        \end{subfigure}
        \begin{subfigure}[b]{0.2660\textwidth}
                \includegraphics[width=1\textwidth]{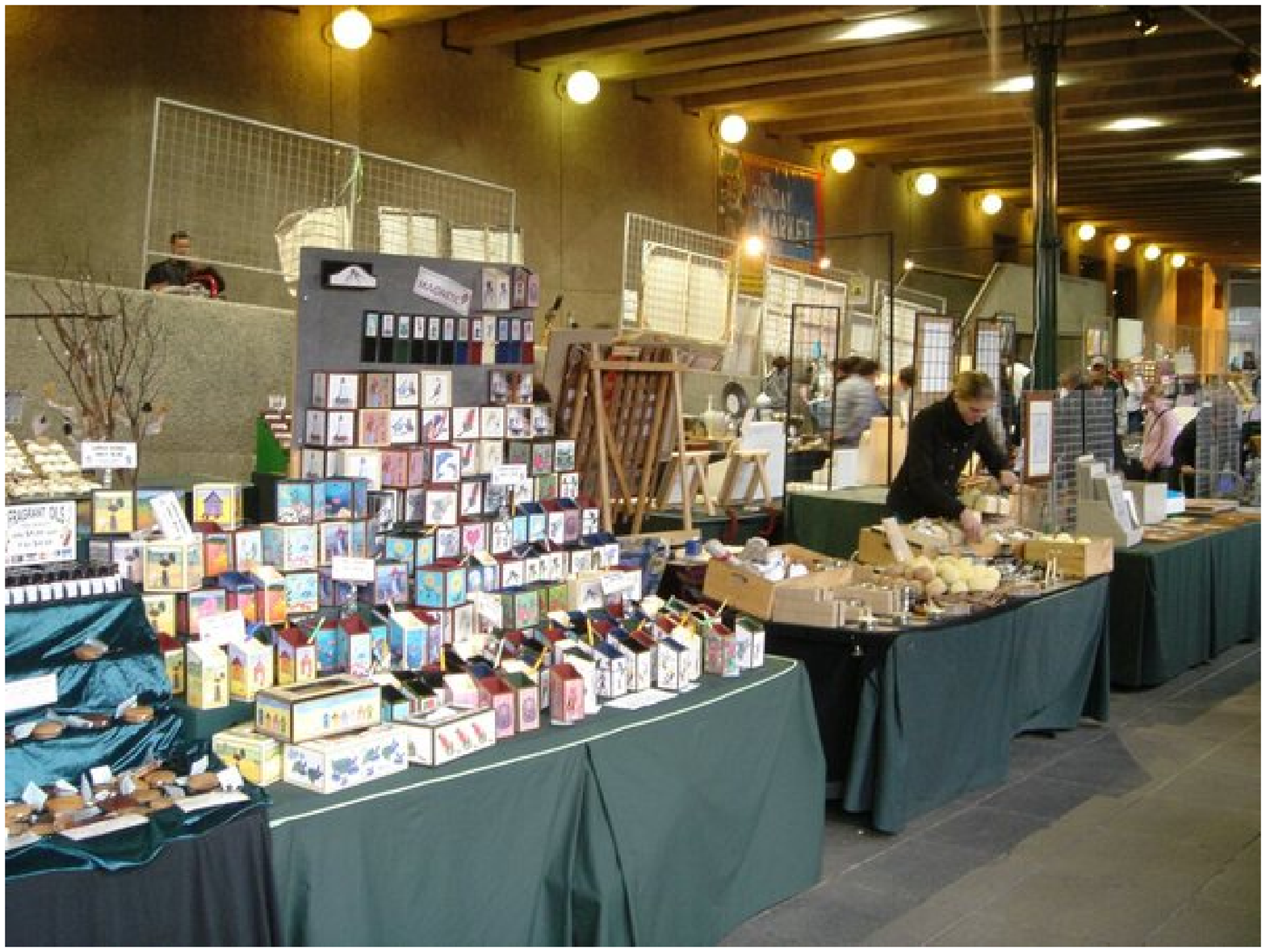}
				\caption{market (indoor)}
        \end{subfigure}
        \begin{subfigure}[b]{0.2660\textwidth}
                \includegraphics[width=1\textwidth]{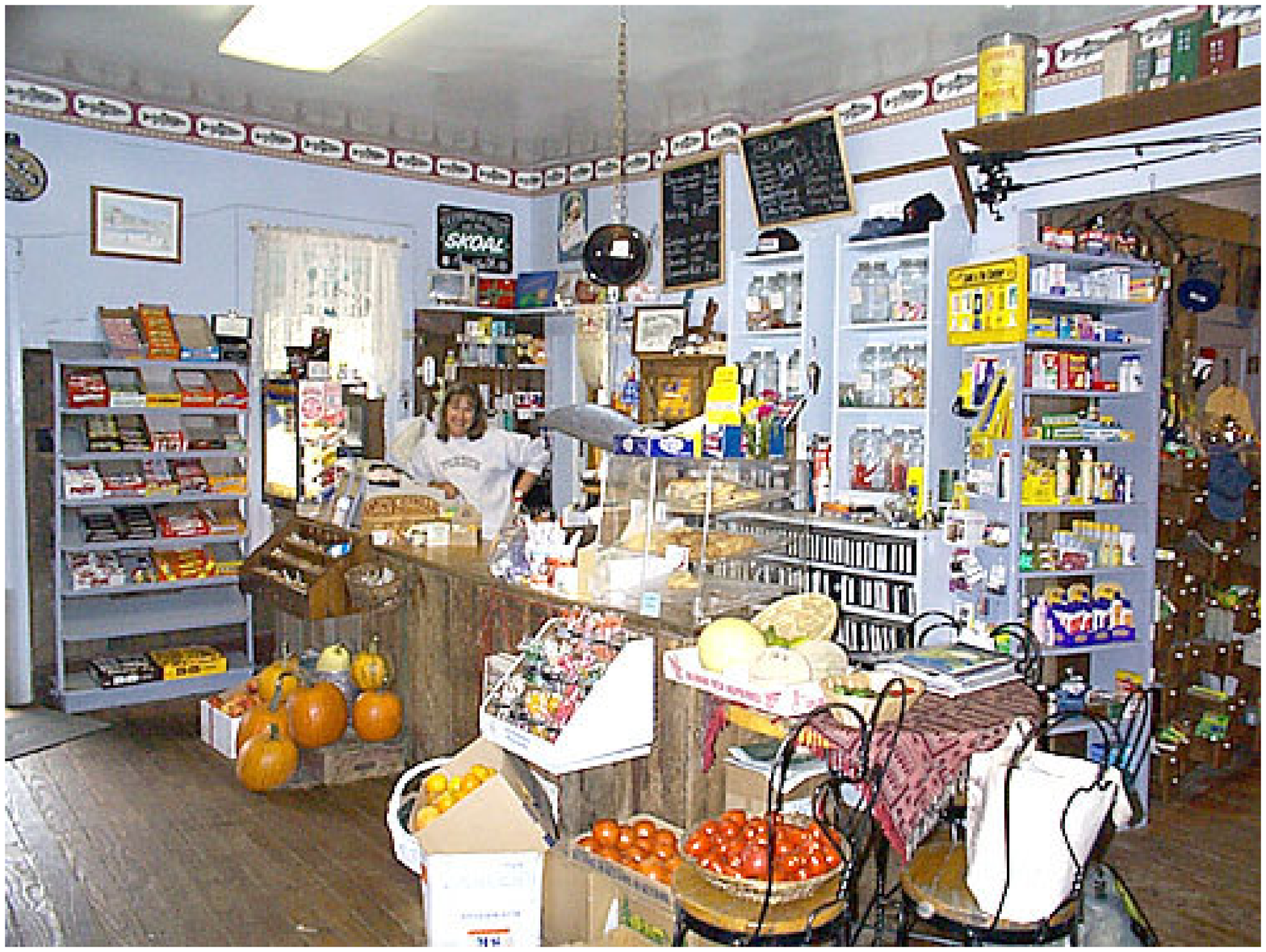}
				\caption{general store (indoor)}
        \end{subfigure}
        \caption{The instance images from four nearest neighboring seen categories of the unseen category ``flea market" with the attributes ``cloth" and ``cluttered space".}
        \label{fig:neighbors}
        \vspace{-10pt}
\end{figure*}

\setlength{\textfloatsep}{5pt}

\noindent\textbf{Qualitative Analysis of the Attributes Learnt by AEZSL:} In order to show the advantage of learning category-specific visual-semantic mappings intuitively, we take the SUN dataset as an example to investigate why our AEZSL can correctly classify the test instances which are misclassified by ESZSL. As illustrated in Figure~\ref{fig:showcases}, the two images are correctly classified as ``flea market" by our AEZSL method but are misclassified as ``shoe shop" by ESZSL. Based on the semantic representations of ``flea market" and ``shoe shop", we find that among the five attributes with maximum values in the semantic representation of ``flea market", only two attributes (\ie, ``cloth" and ``cluttered space") have larger values than those of ``shoe shop". We simply treat these two attributes as the representative attributes to distinguish ``flea market" from ``shoe shop". Figure~\ref{fig:showcases} shows the mapped values of the above two confusing images corresponding to two representative attributes obtained by using ESZSL and our AEZSL method, which are calculated by using ${\X^s}'\W$ and ${\X^s}'\W^c$ ($\W^c$ is the category-specific mapping matrix for ``flea market") respectively. It can be seen that our AEZSL method obtains larger mapped values for the two representative attributes than ESZSL, which contributes to the correct classification of two confusing images as ``flea market".

Based on the fact that AEZSL obtains larger mapped values corresponding to the two representative attributes, we conjecture that the learnt mapping using our method can better capture the semantic meanings of ``cloth" and ``cluttered space". To verify this point, we first show some instance images from different categories with the attribute ``cloth" (\resp, ``cluttered space") in the first (\resp, second) row in Figure~\ref{fig:attributes}. Together with the instance images in Figure~\ref{fig:showcases}, we can observe that for different categories, the visual appearances of ``cloth" and ``cluttered space" are considerably different as discussed in Section~\ref{sec:intro}. Thus, a general mapping matrix learnt by ESZSL cannot tell the subtle discrepancy in the semantic meanings of the same attribute between different categories. In contrast, our AEZSL method learns the category-specific mapping matrix for ``flea market" based on the similarities between this unseen category and each seen category, in which the major transfer comes from more similar seen categories. In Figure~\ref{fig:neighbors}, we show the instance images from four nearest neighboring seen categories of ``flea market" with the attributes ``cloth" and ``cluttered space", from which it can be seen that in terms of the visual appearances of ``cloth" and ``cluttered space", these neighboring categories resemble  ``flea market" much better than other categories such as those in Figure~\ref{fig:attributes}. Therefore, AEZSL can learn a better fitting visual-semantic mapping for ``flea market". We have similar observations for the other unseen categories and on the other datasets.

\setlength{\textfloatsep}{5pt}
\begin{table*}[t]
\caption{Accuracies (\%) of different baseline methods and our DAEZSL method on the ImageNet dataset. The best results are highlighted in boldface.}
\setlength{\tabcolsep}{5pt}
\label{tab:exp_ZSL_large}
\centering
\begin{tabular}{|c|c|c|c|c|c|c|c|c|c|c|}
\hline
\multirow{2}{*}{Setting} & \multirow{2}{*}{Method} & \multicolumn{5}{c|}{Flat Hit@K} & \multicolumn{4}{c|}{Hierarchical precision@K}\\
\cline{3-11}
 & & 1 & 2 & 5 & 10 & 20 & 2 & 5 & 10 & 20 \\
\hline
\multirow{5}{*}{2-hop test} & DeVISE~\cite{frome2013devise} & 6.0 & 10.0 & 18.1 & 26.4 & 36.4 & 15.2 & 19.2 & 21.7 & 23.3\\ 
 & ConSE~\cite{norouzi2013zero} & 9.4 & 15.1 & 24.7 & 32.7 & 41.8 & 21.4 & 24.7 & 26.9 & 28.4\\ 
 & Changpinyo~\etal~\cite{changpinyo2016synthesized} & 10.5 & 16.7 & 28.6 & 40.1 & 52.0 & 25.1 & 27.7 & 30.3 & 32.1\\
 & EXEM~\cite{changpinyo2016predicting} & 12.5 & 19.5 & 32.3 & 43.7 & 55.2 & 26.9 & 29.1 & 31.2 & 33.3 \\
 & ESZSL & 8.6 & 13.8 & 24.1 & 35.0 & 47.4 & 21.8 & 24.7 & 27.4 & 30.2 \\
 & DAEZSL & \textbf{13.9} & \textbf{21.3} & \textbf{33.8} & \textbf{45.2} & \textbf{57.1} & \textbf{28.4} & \textbf{29.8} & \textbf{32.5} & \textbf{34.8} \\
\hline
\multirow{5}{*}{3-hop test} & DeVISE~\cite{frome2013devise} & 1.7 & 2.9 & 5.3 & 8.2 & 12.5 & 3.7 & 19.1 & 21.4 & 23.6 \\
& ConSE~\cite{norouzi2013zero} & 2.7 & 4.4 & 7.8 & 11.5 & 16.1 & 5.3 & 20.2 & 22.4 & 24.7 \\
& Changpinyo~\etal~\cite{changpinyo2016synthesized} & 2.9 & 4.9 & 9.2 & 14.2 & 20.9 & 8.0 & 23.7 & 26.4 & 28.6 \\
& EXEM~\cite{changpinyo2016predicting} & 3.6 & 5.9 & 10.7 & 16.1 & 23.1 & \textbf{8.2} & 25.3 & 27.8 & 30.1 \\
 & ESZSL & 2.4 & 4.1 & 7.6& 11.8 & 17.5& 5.8& 20.9 & 23.1 & 25.2 \\
 & DAEZSL & \textbf{4.3} & \textbf{6.6} & \textbf{11.9} & \textbf{17.6} & \textbf{24.5} & 8.0 & \textbf{26.7} & \textbf{28.9} & \textbf{31.8} \\
\hline
\multirow{5}{*}{``all" test} & DeVISE~\cite{frome2013devise} & 0.8 & 1.4 & 2.5 & 3.9 & 6.0 & 1.7 & 7.2 & 8.5 & 9.6\\
& ConSE~\cite{norouzi2013zero} & 1.4 & 2.2 & 3.9 & 5.8 & 8.3 & 2.5 & 7.8 & 9.2 & 10.4\\
& Changpinyo~\etal~\cite{changpinyo2016synthesized} & 1.5 & 2.4 & 4.5 & 7.1 & 10.9 & 3.6 & 9.6 & 11.0 & 12.5\\
& EXEM~\cite{changpinyo2016predicting} & 1.8 & 2.9 & 5.3 & 8.2 & 12.2 & \textbf{3.7} & 10.4 & 12.1 & 13.5 \\
& ESZSL & 1.2 & 2.0 & 3.8& 5.9& 9.1 & 2.8 & 9.3 & 10.7 & 11.9 \\
 & DAEZSL & \textbf{2.0}& \textbf{3.2}& \textbf{5.9} &\textbf{8.7} & \textbf{13.6} & 3.5 & \textbf{11.8} & \textbf{12.9} & \textbf{14.7} \\
\hline
\end{tabular}
\end{table*}

\noindent\textbf{Performance Variation \wrt Iteration in AEZSL\_LR:} Since our proposed label refinement method is a progressive approach, we are interested in the variation of the accuracy of the predicted test labels (\ie, the union of $\Y^l$ and $\Y^u$) \wrt the number of iterations. By taking the Dogs dataset as an example, we plot the label accuracy \wrt the number of iterations in the left subfigure in Figure~\ref{fig:progressive_refinement}, from which we can observe that the accuracy of predicted test labels increases from 44.62\% to 50.62\% steadily within $50$ iterations. Recall that in each iteration, we move the top $k$ confident instances from the unconfident set with their refined predicted labels into the confident set. Thus, we can infer that in most iterations, the accuracy of the predicted labels of the selected $k$ confident instances is improved using the updated visual classifiers, which verifies the effectiveness of progressive label refinement. We have similar observations on the other datasets.

\noindent\textbf{Generalized ZSL:} 
Most of existing ZSL methods assume that in the testing stage, the test instances only come from the unseen categories, which is actually an unrealistic setting because the instances from seen categories may also be encountered in the testing stage. So it is more useful to predict a test instance from either seen categories or unseen categories instead of assuming that the test instances are only from unseen categories. However, when using the mixture of test instances from both seen and unseen categories for testing, the performance will be significantly degraded due to the bias of prediction scores between seen category label space and unseen category label space, as demonstrated in~\cite{socher2013zero,chao2016empirical}. This more challenging test setting is referred to as generalized zero-shot learning (GZSL) in~\cite{chao2016empirical}. To validate the effectiveness of our AEZSL method under the more realistic GZSL setting, we additionally conduct experiments by mixing the test instances from both seen and unseen categories as the test set on the CUB dataset, which is the overlapped dataset with those used in~\cite{chao2016empirical}. In particular, we follow the setting in~\cite{chao2016empirical}, that is, we move $20\%$ of 
the instances from each of $150$ seen categories to the test set, and thus the test set which originally has $50$ unseen categories is expanded to $200$ test categories including both seen categories and unseen categories. 

When applying our AEZSL method to generalized ZSL setting, we adopt the same strategy as in (\ref{eqn:ESZSL_category}) and the only difference is that we learn $(C^s+C^t)$ instead of $C^t$ category-specific visual-semantic mappings.
Particularly, we learn category-specific mappings for both unseen and seen categories by assigning different weights on different classification tasks of different seen categories based on the similarity between each category and all the seen categories. For fair comparison with state-of-the-art results under the generalized ZSL setting, we further employ the existing calibrated stacking strategy proposed in~\cite{chao2016empirical} on the results obtained by our AEZSL method. The idea of calibrated stacking strategy is simple yet very effective, that is, to reduce the prediction scores in the seen category label space by a threshold, which is learnt based on a performance metric called Area Under SeenUnseen
accuracy Curve (AUSUC) on the validation set, according to the observation that the prediction scores in the seen category label space are often higher than those in the unseen category label space. Thus, after employing the calibrated stacking strategy, we expect to obtain unbiased prediction scores.

\setlength{\textfloatsep}{5pt}
\begin{table}[t]
\caption{Accuracies (\%) of different methods under the generalized ZSL setting on the CUB dataset. The best results are highlighted in boldface.}
\setlength{\tabcolsep}{5pt}
\label{tab:exp_GZSL}
\centering
\begin{tabular}{|c|c|c|c|c|}
\hline
Method & ESZSL & AEZSL & AEZSL(CS) &  Chao~\etal~\cite{chao2016empirical}\\
\hline
Accuracy(\%) & 26.35 & 30.76 & \textbf{42.08} & 35.60 \\
\hline
\end{tabular}
\end{table}

In Table~\ref{tab:exp_GZSL}, we report the results obtained by ESZSL and our AEZSL method as well as our AEZSL after employing Calibrated Stacking (CS) strategy, which is referred to as AEZSL(CS). We also include the state-of-the-art result reported in~\cite{chao2016empirical} for comparison. From Table~\ref{tab:exp_GZSL}, we observe that our AEZSL method outperforms ESZSL, which shows that our AEZSL method is also effective under the generalized ZSL setting. Moreover, after employing the calibrated stacking strategy, the accuracy obtained by our AEZSL method is greatly improved, which demonstrates that our AEZSL method can be perfectly integrated with the existing calibrated stacking strategy. Finally, we observe that AEZSL(CS) achieves significantly better result $42.08\%$ than the state-of-the-art result $35.60\%$ reported in~\cite{chao2016empirical}, which again indicates the effectiveness of our method under the generalized ZSL setting.

\subsection{Zero-Shot Learning on Large-Scale Dataset} \label{sec:exp_large} In this section, we apply our deep adaptive embedding model DAEZSL on the ImageNet dataset and compare with the state-of-the-art reported results.

\noindent\textbf{Experimental Settings:}
We strictly follow the experimental settings in~\cite{frome2013devise,changpinyo2016synthesized}. Specifically,
we use the $1000$ categories in ImageNet ILSVRC 2012
1K ~\cite{russakovsky2015imagenet} as seen categories, and perform evaluation on three test scenarios, which are chosen from  ImageNet 2011 21K dataset and built based on ImageNet label hierarchy with increasing difficulty. The three test sets are listed as follows.

\begin{itemize}
\item 2-hop test set: $1,509$ unseen categories within
two tree hops of the $1000$ seen categories according to
the ImageNet label hierarchy\footnote{http://www.image-net.org/api/xml/structure\_released.xml}, which are semantically and visually similar with $1000$ seen categories.
\item 3-hop test set: $7,678$ unseen categories within three tree hops of $1000$ seen categories, which is constructed in a similar way to 2-hop test set.
\item ``all" test set: all the 20,345 unseen categories in the ImageNet 2011 21K dataset which do not belong to the ILSVRC 2012 1K dataset.
\end{itemize}

Note that 2-hop (\resp, 3-hop) test set is a subset of 3-hop (\resp, ``all") test set, and all the test sets have no overlap with the training set (\eg, ImageNet ILSVRC 2012 1K).

\noindent\textbf{Semantic Representations:}
For both seen categories and unseen categories, we use the  $500$-dim word vector for each category provided in~\cite{changpinyo2016synthesized}, which is obtained based on a skip-gram language model~\cite{mikolov2013distributed} trained on the latest Wikipedia corpus with more than 3 billion words. Note that one ImageNet category may have more than one word according to its synset, we simply average the word vectors of all the words appearing in its synset as the word vector for that category.

\noindent\textbf{Evaluation Metrics:}
Evaluating ZSL methods on the large-scale ImageNet dataset is a nontrivial task considering the large number of unseen categories and the semantic overlap of different categories in the ImageNet label hierarchy. So we use different evaluation metrics compared with multi-class accuracy used on three small-scale datasets (\ie, CUB, SUN, and Dog) in Section~\ref{sec:exp_standard}.
Following \cite{frome2013devise}, we use two metrics Flat hit@K and Hierarchical precision@K for performance evaluation. To be exact, Flat hit@K is defined as the percentage of test instances for which the ground-truth category is in its top $K$ predictions. When $K=1$, Flat hit@K is identical with the multi-class accuracy. Compared with Flat hit@K, Hierarchical precision@K takes the hierarchical structure of categories into consideration. Given each test image, we generate a feasible set of $K$ nearest categories of its ground-truth category in the ImageNet label hierarchy and calculate the overlap ratio between the feasible set and the top $K$ predictions of this test image, that is, the precision of top $K$ predictions. When generating a feasible set of $K$ nearest categories for a ground-truth category, we enlarge the searching radius around the ground-truth category in the ImageNet label hierarchy iteratively and add the searched categories belonging to the test set to the feasible set. This procedure is repeated until the size of feasible set exceeds $K$. For more details, please see the Appendix of~\cite{frome2013devise} or the Supplementary of~\cite{changpinyo2016synthesized}. Note that when $K=1$, Flat hit@K is equal to Hierarchical precision@K, so we omit Hierarchical precision@1 in Table~\ref{tab:exp_ZSL_large} to avoid redundancy.

\noindent\textbf{Network Structure:}
In terms of CNN and MLP in Fig.~\ref{fig:DAEZSL}, we use GoogleNet  as the CNN structure in our experiments, which is initialized with released model~\cite{szegedy2015going} pretrained on ImageNet dataset. The dimension of output from CNN is $1,024$. 
The MLP we use has one hidden layer with its size empirically set as the average of feature dimension and attribute dimension, \ie, $\lfloor \frac{d+a}{2}\rfloor$. Besides, we add a dropout layer following the hidden layer in MLP with $50\%$ ratio of dropped output. Additionally, we add a sigmoid layer following MLP to normalize the feature mask in the range of $(0,1)$.
The network is implemented using TensorFlow and we use AdaGrad optimizer for training, with batchsize as $128$ and learning rate as $0.001$.

\noindent\textbf{Experimental Results:} To the best of our knowledge, there are few ZSL papers \cite{frome2013devise,norouzi2013zero,changpinyo2016synthesized,changpinyo2016predicting} reporting their performances on the ImageNet dataset in 2-hop, 3-hop, and ``all" test settings. We compare our DAESL method with the reported results of DeVISE~\cite{frome2013devise}, ConSE~\cite{norouzi2013zero}, EXEM~\cite{changpinyo2016predicting}, and \cite{changpinyo2016synthesized} in Table~\ref{tab:exp_ZSL_large}. We also include ESZSL as a baseline, in which we train the DAEZSL network using all-one feature masks without learning MLP. From Table~\ref{tab:exp_ZSL_large}, we can observe that DAEZSL achieves far better results than ESZSL, which demonstrates the advantage of category-specific feature mask. Our DAEZSL also outperforms all the baseline methods in most cases (25 out of 27), indicating that it is beneficial to learn deep adaptive embedding model which can implicitly adapt visual-semantic mapping to different categories by using category-specific feature mask. 

\section{Conclusion} \label{sec:conclusion}
In this paper, we have proposed our AEZSL method, which aims to learn a category-specific visual-semantic mapping for each unseen category based on the similarities between unseen categories and seen categories, followed by progressive label refinement. Moreover, we additionally propose a deep adaptive embedding model named DAEZSL for large-scale ZSL, which only needs to be trained once on the seen categories but can be applied to arbitrary number of unseen categories. Comprehensive experimental results demonstrate the effectiveness of our proposed methods.

% You can push biographies down or up by placing
% a \vfill before or after them. The appropriate
% use of \vfill depends on what kind of text is
% on the last page and whether or not the columns
% are being equalized.

%\vfill

% Can be used to pull up biographies so that the bottom of the last one
% is flush with the other column.
%\enlargethispage{-5in}

\bibliographystyle{IEEEtran}
\bibliography{egbib}

% that's all folks
\end{document}